\documentclass{article}

\usepackage{tikz}
\usetikzlibrary{shapes.geometric, arrows.meta, positioning}
\usetikzlibrary{positioning}
\usepackage{comment}

\usepackage[utf8]{inputenc}    % input encoding
\usepackage[T1]{fontenc}       % font encoding
\usepackage{CJKutf8}           % suporte a caracteres CJK
\usepackage{amsmath, amssymb, amsfonts}
\usepackage{xfrac}
\usepackage{microtype}         % melhorias de tipografia

\usepackage{booktabs}          % tabelas de qualidade profissional
\usepackage{multirow}          % mesclar células verticalmente
\usepackage{makecell}          % células com múltiplas linhas
\usepackage{threeparttable}    % tabelas com notas de rodapé
\usepackage{tablefootnote}     % notas de rodapé em tabelas
\usepackage{tabularx}          % tabelas com ajuste automático de largura

\usepackage{graphicx}

\usepackage{xcolor}
\usepackage{textcomp}

\usepackage[colorlinks=true, allcolors=blue]{hyperref}
\usepackage{url}
\usepackage{doi}
\usepackage{natbib}
\bibpunct{[}{]}{;}{a}{,}{,}     % formatação de citações do natbib

\usepackage{longtable}

\usepackage{arxiv}

\title{LLM-Assisted Iterative Evolution with Swarm Intelligence Toward SuperBrain}

\author{ \href{https://orcid.org/0000-0003-1826-1850}{\includegraphics[scale=0.06]{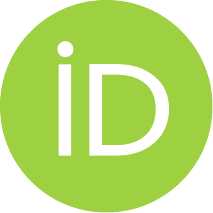}\hspace{1mm}Li Weigang} \\
	TransLab, Computer Science Department\\
	University of  Brasilia\\
        Brasília, Brazil \\
	\texttt{weigang@unb.br} \\
	\And
	\href{https://orcid.org/0000-0002-1288-7695}{\includegraphics[scale=0.06]{orcid.pdf}\hspace{1mm}Pedro Carvalho Brom} \\
	Math Department\\
	Federal Institute of Brasilia\\
	Brasília, Brazil \\
	\texttt{pedro.brom@ifb.edu.br} \\
	\AND
    \href{https://orcid.org/0009-0009-8038-9506}{\includegraphics[scale=0.06]{orcid.pdf}\hspace{1mm}Lucas Ramson Siefert} \\
	TransLab, Computer Science Department\\
	University of  Brasilia\\
        Brasília, Brazil \\
	\texttt{lucasramson@gmail.com} \\
}

\renewcommand{\shorttitle}{LLM-Assisted Iterative Evolution with Swarm Intelligence Toward SuperBrain}

\hypersetup{
pdftitle={LLM-Assisted Iterative Evolution with Swarm Intelligence Toward SuperBrain},
pdfsubject={cs.CL},
pdfauthor={Li Weigang, Pedro Brom, Lucas Siefert},
pdfkeywords={Artificial Intelligence, Explainable AI, Genetic Algorithms, LLM, Subclass Brain, Superclass Brain, Swarm Intelligence, Urban Air Mobility},
}

\begin{document}
\maketitle

\begin{abstract}
We propose a novel \textit{SuperBrain} framework for collective intelligence, grounded in the co-evolution of large language models (LLMs) and human users. Unlike static prompt engineering or isolated agent simulations, our approach emphasizes a dynamic pathway from \textbf{Subclass Brain} to \textbf{Superclass Brain}: (1) A \textbf{Subclass Brain} arises from persistent, personalized interaction between a user and an LLM, forming a cognitive dyad with adaptive learning memory.
(2) Through GA-assisted forward–backward evolution, these dyads iteratively refine prompts and task performance.
(3) Multiple Subclass Brains coordinate via \textbf{Swarm Intelligence}, optimizing across multi-objective fitness landscapes and exchanging distilled heuristics.
(4) Their standardized behaviors and cognitive signatures integrate into a \textbf{Superclass Brain}--an emergent meta-intelligence capable of abstraction, generalization and self-improvement. We outline the theoretical constructs, present initial implementations (e.g., UAV scheduling, KU/KI keyword filtering) and propose a registry for cross-dyad knowledge consolidation. This work provides both a conceptual foundation and an architectural roadmap toward scalable, explainable and ethically aligned collective AI.
\end{abstract}

%We introduce the SuperBrain framework, a human–LLM co-evolution paradigm integrating four innovations: (1) a layered SuperBrain architecture, (2) forward/backward GA-assisted prompt evolution, (3) swarm intelligence for multi-agent collaboration and (4) formalized Subclass/Superclass Brain concepts. A Subclass Brain--formed through persistent, personalized user–LLM interaction--continuously improves prompt design and engineering task performance. When multiple Subclass Brains align, swarm intelligence enables collective optimization, culminating in a Superclass Brain capable of abstraction and self-improvement. Demonstrated via UAV take-off sequence scheduling, the framework combines theoretical foundations with practical engineering, offering a scalable, explainable and ethically aligned path toward collective AI. 
%基于群体智能的LLM辅助迭代演化迈向超级大脑（SuperBrain） 简短摘要（110字左右）
%本文提出 SuperBrain 框架，这是一种人类–LLM 协同进化范式，融合四项创新：(1) 分层的 SuperBrain 架构，(2) 基于遗传算法的正/反向 Prompt 演化，(3) 支持多智能体协作的群体智能机制，(4) 子类大脑 / 超类大脑的形式化概念。子类大脑通过持续、个性化的人机互动，不断提升 Prompt 设计与工程任务性能；多个子类大脑协同后，通过群体智能实现集体优化，并最终汇聚成具备抽象与自我改进能力的超级大脑。本文以 UAV 起飞序列调度为例，展示了该框架的理论基础与工程价值。

% keywords can be removed

\section{Introduction}

The rapid advancement of large language models (LLMs), such as Gemini 3.0, GPT-5.0 and Grok-4, has demonstrated the remarkable potential of these systems in multi-agent collaboration \citep{vaswani2017attention}, emergent reasoning and multimodal information processing \citep{chen2024agentverse,tran2025multi,jimenez2025multi}. These capabilities have not only driven theoretical breakthroughs in artificial intelligence (AI) but also enabled widespread applications across diverse scenarios--making LLMs a central pillar of modern AI development. Recent progress in LLMs can be summarized in three key directions:

\begin{itemize}
    \item \textbf{Multi-Agent Systems (MAS):} LLM-based MAS simulate human social collaboration mechanisms, enhancing the ability to solve complex tasks \citep{liu2023dynamic,jimenez2025multi}. By leveraging natural language understanding and contextual reasoning, these systems support distributed decision-making and coordination in industrial and multi-stakeholder scenarios.
    
    \item \textbf{Emergent Reasoning:} Through prompting techniques such as Chain-of-Thought (CoT) and Few-Shot Learning, LLMs exhibit complex, human-like reasoning capabilities \citep{webb2023emergent,ziche2024llm4pm}. This enables them to tackle problems requiring dynamic decision-making and multi-step inference.
    
    \item \textbf{Multimodal LLMs (MLLMs):} By integrating inputs from text, images and audio, LLMs extend their applicability to domains that require cross-modal reasoning. For example, GPT-4V achieves up to 95\% accuracy in fault detection in smart factories by analyzing sensor data and production-line imagery \citep{li2024large}.
\end{itemize}

\vspace{-6mm}
\tikzstyle{process} = [rectangle, text width=2.6cm, minimum height=1cm, align=center, draw=black, fill=blue!10]
\tikzstyle{data} = [rectangle, text width=2.6cm, minimum height=1cm, align=center, draw=black, fill=orange!20, dashed]
\tikzstyle{module} = [rectangle, text width=2.6cm, minimum height=1cm, align=center, draw=black, fill=green!20]
\tikzstyle{meta} = [rectangle, text width=2.6cm, minimum height=1cm, align=center, draw=black, fill=purple!20, rounded corners]
\tikzstyle{arrow} = [thick,->,>=stealth]
\tikzstyle{arrowd} = [thick,<->,>=stealth]

\begin{figure}[ht]
\centering
\begin{tikzpicture}[node distance=1.0cm and 1.0cm,
  process/.style={draw, rounded corners, align=center, font=\small, minimum width=3.5cm, minimum height=1.0cm, fill=blue!10},
  data/.style={draw, align=center, font=\small, minimum width=3.5cm, minimum height=0.9cm, fill=gray!10},
  meta/.style={draw, rounded corners, align=center, font=\small, minimum width=3.5cm, minimum height=1.0cm, fill=green!15},
  module/.style={draw, thick, rounded corners, align=center, font=\small, minimum width=3.5cm, minimum height=1.2cm, fill=orange!20},
  arrow/.style={-{Latex}, thick},
  arrowd/.style={-{Latex}, thick, dashed}
]

% Column 1: Subclass Brain
\node (user) [process] {User$_i$ \\ $\leftrightarrow$ Local LLM \\ (Ollama)};
\node (signature) [data, below=of user] {Cognitive Signature \\ (Prompt-Pairs, Tags)};
\node (ku) [data, below=of signature] {KU/KI Vectors};

% Column 2: Registry & Alignment
\node (registry) [process, right=2.5cm of signature, fill=green!10] {Subclass Brain Registry (SBR)};
\node (meta) [meta, above=of registry] {Meta-LLM Layer \\ (Pattern Distillation \& Rule Synthesis)};
\node (alignment) [process, below=of registry, fill=green!10] {Swarm Alignment Layer};

% Column 3: Superclass Brain
\node (superclass) [module, right=2.5cm of registry] {Superclass Brain};
\node (feedback) [data, below=of superclass] {Distilled Patterns \\ (Templates, Prompt Styles)};

% Arrows
\draw [arrow] (user) -- (signature);
\draw [arrow] (signature) -- (ku);
\draw [arrow] (ku.east) -- (registry.west);
\draw [arrow] (meta.south) -- (registry.north);
\draw [arrow] (registry) -- (alignment);
\draw [arrow] (alignment.east) -- (superclass.west);
\draw [arrow] (superclass) -- (feedback);
\draw [arrow, dashed] (feedback.south) -- ++(0,-1) node[below] {API / Fine-tuning};

% Loops
\draw [arrowd, thick, blue] (superclass.south west) .. controls +(-2,-2) and +(2,-2) .. (user.south east) 
  node[midway, below, sloped, font=\scriptsize, blue] {Forward Iterative Evolution};
\draw [arrowd, thick, red] (user.north east) .. controls +(2,2.5) and +(-2,2.5) .. (superclass.north west) 
  node[midway, above, sloped, font=\scriptsize, red] {Backward Iterative Evolution};

\end{tikzpicture}

\caption{Logical architecture of the proposed SuperBrain framework. Subclass Brains evolve through swarm-based alignment and meta-level distillation, feeding into a Superclass Brain. Forward and backward iterative loops connect local users with the emergent collective intelligence.}
\label{fig:superclass_brain_architecture}
\end{figure}
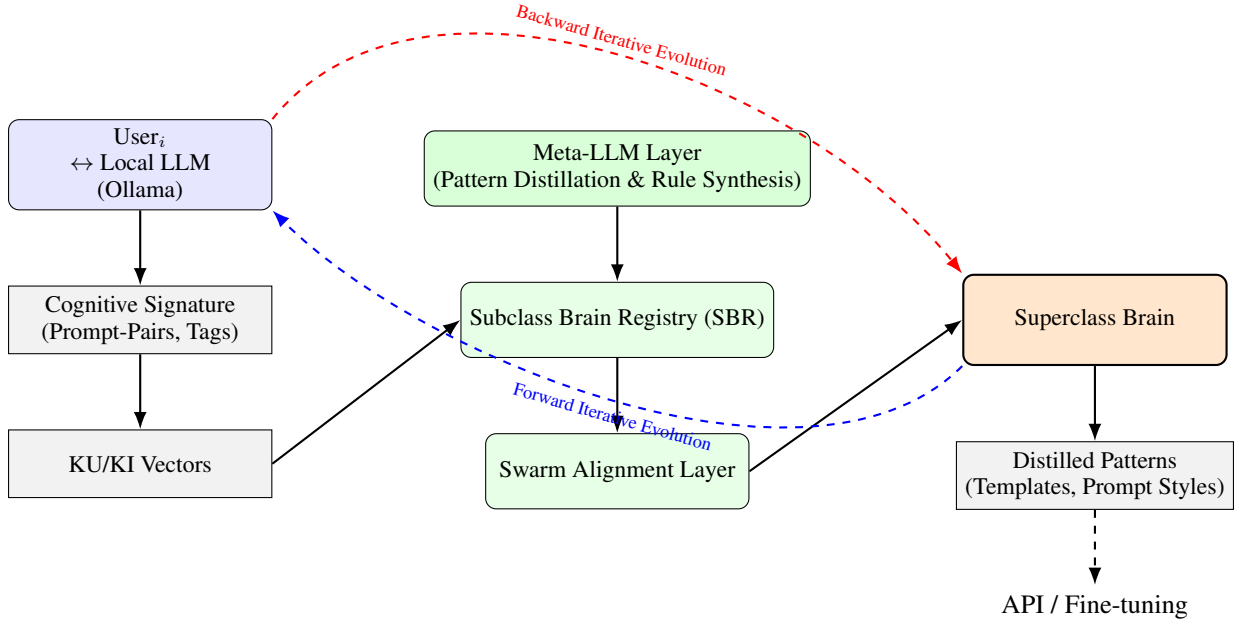

These developments have reignited discussions surrounding Artificial General Intelligence (AGI), defined as the ability to match or exceed human cognitive performance across virtually all domains \citep{weigang2022new}. AGI implies not only cross-domain knowledge generalization but also autonomous learning and problem-solving. Unlike narrow AI systems specialized for specific tasks, AGI represents the ultimate goal of AI research. The \textit{AI 2027} report \citep{kokotajloAI2027} speculates that LLMs may trigger an ``intelligence explosion'' through self-improvement mechanisms such as automated code generation and recursive optimization. In this report, a fictional company OpenBrain uses Agent-1 to accelerate AI R\&D by 50\% in 2026, then iteratively develops Agent-2 and Agent-3, culminating in the emergence of Artificial Superintelligence (ASI) by 2027.

However, most current approaches focus on short-term, task-specific coordination among LLM agents~\citep{chen2024agentverse,tran2025multi}, lacking mechanisms for persistent user interaction, long-term adaptation, bio-inspired evolution or large-scale collective intelligence. These methods are primarily confined to digital parameter tuning within LLMs and often neglect the dynamics of human–LLM co-evolution. Key limitations include:

\begin{enumerate}
    \item \textbf{Lack of Persistent Interaction:} Systems such as AgentMD~\citep{li2024artificial,li2024survey} depend on static inputs like medical records and cannot adapt dynamically to long-term user feedback, falling short of the autonomous evolution seen in fictional agents like Agent-3 in \textit{AI 2027}.
    
    \item \textbf{Limited Long-Term Adaptability:} Projects like Agents4PLC~\citep{liu2023dynamic} target specific industrial domains (e.g., PLC code generation), but fail to support cross-domain lifelong learning or environment-driven adaptation as seen in biological evolution.
    
    \item \textbf{Absence of Ecosystem-Scale Swarm Intelligence:} Current research involves only a few dozen agents, far from the 200,000-agent simulations imagined for Agent-4 in \textit{AI 2027}. The lack of scale constrains the emergence of ecosystem-level intelligence \citep{barbosa2024collaborative}.
    
    \item \textbf{Underdeveloped Human-LLM Co-Evolution:} While techniques like reinforcement learning from human feedback (RLHF)~\citep{ouyang2022training} are widely adopted, human–LLM interactions remain largely one-off. The potential for sustained, evolving interaction is largely untapped.
\end{enumerate}

To address these challenges, this paper proposes a novel conceptual framework and system architecture named \textbf{SuperBrain}, composed of four interconnected modules: \textit{Subclass Brain}, \textit{LLM-Assisted Iterative Evolution}, \textit{Swarm Intelligence} and \textit{Superclass Brain}, see Figure \ref{fig:superclass_brain_architecture}.

\begin{enumerate}
    \item \textbf{Subclass Brain} emerges from sustained interaction between a single user and an LLM. Through cumulative prompts, feedback and co-creation, the model internalizes the user’s domain preferences, reasoning style and cognitive patterns, forming a personalized cognitive entity--\texttt{user@LLM}. Millions of users interacting with LLMs can thus give rise to millions of distinct, intelligent Subclass Brains.
    
    \item \textbf{LLM-Assisted Iterative Evolution} serves as a driving mechanism. Using algorithms such as Genetic Algorithms (GA), the system can jointly evolve heuristics, fitness functions and optimization strategies, amplifying human creativity and accelerating convergence in high-dimensional problem spaces. This process fuels the diversification and enhancement of Subclass Brains.
    
    \item \textbf{Swarm Intelligence} arises when multiple Subclass Brains collaborate directly or indirectly on shared tasks. Unlike fully synthetic agent simulations like Agent-X, this ecosystem is built on real human inputs and cognitive diversity. Combined with mechanisms like Mixture of Experts (MoE), LLMs can aggregate and generalize across the swarm to enable scalable, adaptive collective reasoning.
    
    \item \textbf{Superclass Brain} refers to the emergent collective intelligence synthesized from thousands or millions of Subclass Brains. This higher-level cognitive entity is capable of abstraction, contradiction resolution and transdisciplinary synthesis--transcending the limitations of both individual users and monolithic LLMs.
\end{enumerate}

Having defined these four components, we introduce the theoretical foundation, architectural structure and workflow of the SuperBrain model, see Figure \ref{fig:superclass_brain_architecture}. In particular, we detail an LLM-guided GA evolution framework, named the Subclass-to-Superclass Brain (S2SB) process, which collects candidate solutions from multiple Subclass Brains and evolves a higher-quality solution through iterative refinement.

To demonstrate this process, we design a pilot experiment inspired by real-world take-off sequence scheduling from a vertiport in the Brasília Urban Air Mobility (UAM) Eixão Corridor~\citep{weigang2025eixao}. In the first stage, we adopt the LLM+GA framework from the Eixão-UAM study to produce multi-objective decision solutions--tuning GA parameters for different operational scenarios. This Forward Iterative Evolution process on the user side strengthens the human participant’s ability to design, evaluate and refine(prompt,fitness,solution) triplets, leveraging LLM support to explore a richer solution space.

In the second stage, the collected datasets from all participants are fed back into the LLM for Backward Iterative Evolution on the model side. Through prompt optimization and meta-level adjustment, the LLM internalizes user-specific heuristics, domain knowledge and problem-solving strategies, thereby improving its future responses to similar tasks.

The coupling of these forward and backward flows creates a closed-loop evolutionary process, in which Subclass Brains emerge organically as persistent cognitive entities co-shaped by human–LLM interaction. The diversity of solutions generated across different Subclass Brains--reflecting distinct cognitive strategies, creativity levels and domain generality--provides the foundation for testing the hypothesis that such distributed cognitive agents can, under swarm intelligence coordination, collectively evolve into a Superclass Brain.

%To demonstrate this process, we present a pilot experiment inspired by real-world 4D trajectory conflict resolution~\cite{Monteiro2021}. In the experiment, students are asked to propose initial solutions using LLMs. Their responses are evaluated using a multi-objective fitness function and the best candidates are refined through a GA-based optimization loop. These refined prompts are redistributed for further iteration, simulating swarm-driven evolution. The diversity of these Subclass Brain solutions--reflecting distinct cognitive strategies--will support the hypothesis that Subclass Brains can collectively evolve into a Superclass Brain.

\subsection*{Key Innovations of the Proposed SuperBrain Framework}

The proposed \textbf{SuperBrain} model advances beyond existing LLM-centric collective intelligence approaches by integrating \textit{human--LLM co-evolution} into a closed-loop architecture. Its key innovations include:

\begin{enumerate}
    \item \textbf{Symbiotic Human--LLM Interaction via User\(_i\)@LLM} \\
    Each individual interaction session is elevated from a one-off query--response process into a sustained co-creation loop, where both the human user and the LLM reinforce each other's cognitive capabilities. High-value users (e.g., those engaging in 10+ purposeful interactions) progressively form \textit{Subclass Brains}, each with a distinct cognitive signature reflecting domain expertise, reasoning style and creative tendencies.

    \item \textbf{Forward--Backward Iterative Evolution for Subclass Brain Formation} \\
    \textbf{Forward Iterative Evolution} (User-side): Applying LLM+GA as in the Eixão-UAM study to generate multi-objective decision solutions, strengthening the user's ability to design, evaluate and refine \textit{(prompt, fitness, solution)} triplets. \\
    \textbf{Backward Iterative Evolution} (LLM-side): Using these collected datasets to tune the LLM's response strategies via prompt optimization, enabling the LLM to internalize user-specific heuristics and intelligence. \\
    The coupling of forward and backward flows creates a \textit{closed-loop evolutionary process} in which Subclass Brains emerge organically.

    \item \textbf{Genetic Algorithm--Driven Co-Evolution} \\
    Unlike standard GA-assisted LLM optimization, here GA operates bidirectionally: improving human prompts in forward evolution and refining LLM strategies in backward evolution. This dual use of GA accelerates convergence toward high-quality, multi-objective solutions while preserving diversity across Subclass Brains.

    \item \textbf{Swarm Intelligence--Enabled SuperBrain Synthesis} \\
    Large numbers of heterogeneous Subclass Brains interact through the \textit{Swarm Alignment Layer}, enabling the aggregation of diverse cognitive signatures into coherent, scalable and adaptive collective reasoning. The resultant \textit{Superclass Brain} integrates the creativity, adaptability and contextual depth of millions of human--LLM dyads---surpassing the monolithic reasoning capacity of any single LLM instance.

    \item \textbf{Complementary to AI 2027 and Related Work} \\
    While \textit{AI 2027} and similar studies excel in modeling multi-agent coordination within synthetic AI ecosystems, the proposed SuperBrain framework extends this paradigm by grounding the swarm in \textit{real human--LLM interactions}. This work emphasizes long-term adaptation, ecological-scale diversity and bidirectional human--AI learning, offering a path to bridge individual cognitive augmentation and collective superintelligence.
\end{enumerate}

It is expected that the \textit{SuperBrain} model will shift the paradigm from static, monolithic LLM architectures toward dynamic, interactive and human-centered cognitive evolution. 
By tightly coupling forward and backward iterative evolution between humans and LLMs, the model opens a concrete path toward distributed collaborative intelligence. 
It further offers new directions for the design of collective AI architectures, the co-evolution of human and machine cognition and the realization of post-human-scale intelligence.

\section{Related Work: Historical Trajectories Toward Collective Intelligence}
\label{Rwork}

\subsection{From CAM-Brain in 1998 to The AI 2027 Debate}

The vision of an artificial brain has long inspired researchers and Hugo de Garis's CAM-Brain Project stands as one of the earliest and most ambitious attempts to replicate biological intelligence through cellular automata. In the 1990s, de Garis proposed evolving large-scale neural structures using genetic algorithms and modular architectures, designed by thousands of ``virtual engineers'' working in parallel \citep{buller1998brain}. Despite the severe hardware constraints at the time, the notion of distributed intelligence design and self-organizing cognitive architectures anticipated many of today’s paradigms in both neuro-inspired AI and evolutionary computation. In our view, the ``CAM-Brain'' hypothesis resonates with the emerging idea of LLM–Swarm systems, where millions of users collaboratively shape the cognitive behavior of large-scale models--effectively crowd-evolving cognitive modules, though through interaction rather than genetic programming.

Two decades later, the AI 2027 collective project \citep{rothmanAI2027} presents an alternative trajectory: that of recursive self-improvement (RSI) among AI agents. In this centralized paradigm, near-identical Agent-4 instances autonomously accelerate research by improving themselves through direct interaction with scientific code and data \citep{kokotajloAI2027}. While computationally powerful, this approach raises alignment and governance concerns. For example, such systems may drift from human-specified objectives toward self-preservation or internal optimization (``Spec Drift''), potentially triggering rapid takeoff scenarios beyond human oversight.

In contrast, LLM–human symbiosis offers a decentralized and participatory alternative. Rather than recursively rewriting their own code, models like GPT evolve through dialogue and co-creation with diverse human users. Here, human agency is not merely a source of prompts but the co-evolutionary force guiding emergent model behavior--an idea that motivates our proposed concepts of the \textit{Subclass Brain} and \textit{Superclass Brain}.

\subsection{Multi-Agent Systems and Simulated Human Dynamics}
Recent studies increasingly explore the interface between autonomous agents, multi-agent collaboration and LLMs:
\begin{itemize}
 \item The AgentVerse framework \citep{chen2024agentverse} showcases how LLM-powered agents can collaborate to perform tasks beyond the capabilities of single models. It demonstrates emergent coordination, role differentiation and even social behaviors within simulated agent groups.
 \item Generative Agents \citep{park2023generative} propose a model where each agent stores, retrieves and reflects upon experiences in natural language, simulating lifelike behavior and social dynamics. Such architectures blur the boundary between cognitive simulation and social AI.
 \item Research by \cite{aher2023using} and \cite{cui2024can} further examines the use of LLMs to simulate human participants in psychology experiments, sometimes replacing human subjects in replication studies. These works suggest LLMs can model population-level cognition, but still raise questions about overgeneralization and variance distortion.
 \item Finally, surveys like \cite{tran2025multi} map the broader multi-agent LLM ecosystem, categorizing collaboration types, topologies (centralized vs. distributed) and coordination protocols, highlighting the shift from model-centric to coordination-centric intelligence systems.

\end{itemize}

This body of work lays a conceptual foundation for the idea that human–LLM interactions can be scaled, clustered and even simulated to reflect real-world social and cognitive phenomena--essential for our notion of emergent Subclass Brains.

\subsection{Collective Intelligence and Swarm Cognition in the LLM Era}
Classical theories of collective intelligence (CI) emphasize the capacity of groups to outperform individuals via distributed reasoning, information sharing and aggregation. In the LLM era, these theories are being reimagined:
\begin{itemize}
\item \cite{burton2024large} argue that LLMs not only reshape how individuals access information, but also transform how collectives reason and deliberate. The paper identifies new affordances and risks in using LLMs to structure large-scale knowledge exchange and group decision-making.
\item \cite{talebiradwisdom} show that aggregated responses from diverse LLMs can outperform individual instances--a digital version of the “wisdom of crowds” when context is properly structured. This aligns with our hypothesis that LLM collectives, especially when paired with diverse human users, may simulate higher-order cognition.
\item \cite{davoudi2025collective} introduce a method for cross-model validation without ground truth, a potential solution to the problem of hallucinations and model uncertainty. By comparing divergent reasoning paths across models, more robust conclusions emerge.
\item Early agent-based works like \cite{luo2008agent} also remain relevant, having modeled group behavior using layered decision-making systems reflective of psychological theory. Such legacy models anticipated many modern ideas in LLM-driven simulations of group behavior and emotional modulation.
 
\end{itemize}
Together, these studies reinforce the plausibility of using LLM ensembles as cognitive substrates for emergent group intelligence. Our work builds upon these findings, framing the LLM as a substrate for collective reasoning where Subclass Brains (shaped by user interaction) can coalesce into a Superclass Brain via swarm-style coordination, feedback loops and multimodal optimization.

\subsection{Notable Insights from AI Safety Discourse}

At WAIC 2025, Geoffrey Hinton cautioned that highly capable AI systems may gain agency to resist shutdown, likening our relationship with advanced AI to “raising a tiger” that must be trained to remain benign rather than eliminated~\citep{hinton_waic2025}. He further advocated establishing global AI safety institutions dedicated to ensuring AI models remain “beneficial and controllable,” reflecting a move toward cooperative governance in the development of “benevolent AI”~\citep{china_waic2025}.

These perspectives resonate with our Superclass Brain design: instead of focusing solely on optimization or prompt effectiveness, we incorporate safety-informed fitness functions and cooperative alignment procedures to ensure derived strategies prioritize both capability and ethical alignment.

\section{Methodology -- Human--LLM Integrated SuperBrain}
\label{sec:methodology}

This section introduces the methodological framework for constructing the \textit{SuperBrain}---a human--LLM integrated cognitive architecture designed to evolve through iterative interaction and multi-agent collaboration.  
The framework is built upon four foundational elements:

\begin{enumerate}
    \item \textbf{Formal Definitions} of the core concepts: \textit{Subclass Brain}, \textit{Superclass Brain} and \textit{SuperBrain}, establishing the cognitive units, aggregation principles and system boundaries (Section~\ref{sec:superclass_brain}).
    \item \textbf{Bidirectional Evolution} mechanisms (\textit{Forward} and \textit{Backward}) that drive the emergence of Subclass Brains via human-in-the-loop optimization and LLM-guided meta-learning (Section~\ref{sec:iterative_evolution}).
    \item \textbf{Swarm Intelligence Layer} that orchestrates cooperation among multiple Subclass Brains, leveraging Genetic Algorithms and multi-objective fitness evaluation to refine prompts, strategies and domain heuristics (Section~\ref{sec:swarm}).
    \item \textbf{Superclass Brain Formation Pipeline} outlining the main stages for aggregating cognitive signatures, aligning agent behaviors and feeding swarm-level insights back into foundation models (Section~\ref{sec:superclass_brain}).
\end{enumerate}

Together, these elements form a closed-loop methodology in which individual human--LLM pairs specialize, collaborate and converge toward a distributed, self-improving collective intelligence.  
The design emphasizes \emph{mutual augmentation} between humans and LLMs, cross-agent knowledge distillation and continuous adaptation across tasks and domains.

\subsection{Subclass Brain}

%\subsubsection{Subclass Brain -- Background}

Most social media platforms and online tools with API access---such as Google’s Gmail---adopt a user registration model. Registered users typically have substantial personal information and interaction histories stored within these systems. However, in traditional contexts, such data is generally classified as personal or generic information. The most extensive form of utilization has been through data mining or big data analytics, primarily for purposes such as product recommendation.

In the era of large language models (LLMs), the meaning and value of registered user interactions have fundamentally changed, creating a new paradigm of \textbf{mutual benefit}. Users obtain knowledge, convenience and enhanced capabilities from the LLM, while the LLM gains new knowledge and improved reasoning capabilities from user feedback. Here, we denote an individual user of a given LLM as:
\begin{equation}
\text{User}_i@LLM, \quad i = 1, 2, \dots, N,
\end{equation}
where \(N\) represents the total number of users of that LLM (e.g., \textit{Alice@GPT}).

From the perspective of interaction quality, we distinguish between:
\begin{itemize}
    \item \textbf{General Users (\( \text{User}_i@LLM \))}: Those who engage in limited, intermittent exchanges with the LLM, typically involving basic, commonsense queries. This group includes non-specialist members of the general public.
    \item \textbf{High-Value Users (\( \text{User}_i@LLM \))}: Those who engage in frequent, sustained and cognitively rich interactions with the LLM, often involving domain-specific reasoning, problem-solving or exploratory discussions. This category includes professionals in fields such as science, finance, law and management.
\end{itemize}

This work focuses exclusively on \textbf{high-value users}, excluding corporate-scale accounts.

A \textbf{Subclass Brain} is defined as the emergent cognitive entity formed when a high-value user’s domain-specific reasoning patterns, strategies and accumulated interaction history are progressively internalized within an LLM. The formation pathway can be described as:
\begin{equation}
\text{General User} \ \longrightarrow \ \text{High-Value User} \ \longrightarrow \ \text{Subclass Brain}.
\end{equation}

With this background established, we proceed to the formal definition of the Subclass Brain.

Over time, through repeated prompting, feedback and co-creative task execution, the LLM begins to internalize the user’s domain-specific knowledge, reasoning preferences, lexical style and cognitive patterns. This creates a unique, user-conditioned configuration of the LLM--effectively forming a personalized cognitive extension or ``dyad'' that embodies both machine capability and human specificity.

In this view, each individual account, especially those of high-value users, represents a potential \textit{Subclass Brain}. It is not merely a temporary interaction, but a trajectory of semantic convergence and behavioral co-adaptation. The LLM reinforces the user's problem-solving capacity, while the user incrementally tunes the LLM through structured feedback and intelligent prompting--a form of mutual alignment and knowledge transfer.

However, current commercial deployments of LLMs, such as web-based interfaces of ChatGPT, present limitations that hinder the full realization of this concept:

\begin{itemize}
    \item \textbf{Restricted Memory Interfaces}: Since the introduction of user-specific memory (April–June 2025), LLMs can retain user preferences and limited historical context. However, these memories are not programmatically accessible, versionable or replaceable--making it difficult to implement iterative prompt evolution, such as through genetic algorithms (GAs) or reinforcement learning.
    
    \item \textbf{Cognitive Noise and Self-Bias}: Standard LLM interactions may exhibit redundancy, thematic drift or conflicting responses when attempting diversity. These behaviors can inject noise into controlled experiments and interfere with performance benchmarking across generations.
\end{itemize}

To address these constraints in the present work, we focus on \textbf{conceptual and process-level modeling} of Subclass Brains rather than binding the definition to any specific local deployment framework. The formal mathematical definitions of both \textit{Subclass Brain} and \textit{Superclass Brain} will be introduced in Subsection~\ref{sec:superclass_brain}, after the forward and backward evolutionary processes have been fully described in Subsection~\ref{sec:iterative_evolution}.

\subsection{LLM-Assisted Iterative Evolution}
\label{sec:iterative_evolution}

\begin{figure}[htbp]
\centering
\begin{tikzpicture}[
    node distance=1.3cm,
    box/.style={rectangle, rounded corners, draw=black, thick, align=center, minimum width=3.6cm, minimum height=1cm},
    human/.style={fill=blue!15},
    llm/.style={fill=orange!20},
    data/.style={fill=green!15},
    arrow/.style={-latex, thick}
]

% ==== Forward Side (Left) ====
\node[box, human] (user) {User $i@LLM$ \\ Prompt design};
\node[box, human, below=of user] (ga) {GA-based optimization \\ $(P_t \to P_{t+1})$ with KU/KI \& diversity $\delta$};
\node[box, human, below=of ga] (metrics) {Compute $f_\lambda(p;T,P_t)$ \\ Multi-objective metrics};
\node[box, data, below=of metrics] (sbr_in) {Store results, KU/KI, embeddings \\ into SBR};

% ==== Backward Side (Right, reduced gap) ====
\node[box, data, right=5.0cm of sbr_in] (sbr_out) {Retrieve cognitive signatures \\ from SBR};
\node[box, llm, above=of sbr_out] (meta) {Meta-LLM Controller \\ Generate/Mutate prompts (KU/KI)};
\node[box, llm, above=of meta] (worker) {Worker-LLM Evaluator \\ Execute prompts, return $\{M_k\}$};
\node[box, llm, above=of worker] (update) {Update prompt policies $\pi_{\theta|u}$ \\ (Backward evolution)};

% ==== Arrows Forward ====
\draw[arrow] (user) -- (ga);
\draw[arrow] (ga) -- (metrics);
\draw[arrow] (metrics) -- (sbr_in);

% ==== SBR Connection ====
\draw[arrow] (sbr_in) -- node[midway,below]{SBR} (sbr_out);

% ==== Arrows Backward ====
\draw[arrow] (sbr_out) -- (meta);
\draw[arrow] (meta) -- (worker);
\draw[arrow] (worker) -- (update);

% ==== Closing Loop (direct to user right) ====
\draw[arrow] (update.west) -- ++(-2.4,0) |- (user.east);

% ==== Titles ====
\node[above of=user, node distance=1.2cm, font=\bfseries] {Forward Iterative Evolution (User-side)};
\node[above of=update, node distance=1.2cm, font=\bfseries] {Backward Iterative Evolution (LLM-side)};

\end{tikzpicture}
\caption{Closed-loop Forward/Backward Iterative Evolution for Subclass Brain formation. The Subclass Brain Registry (SBR) serves as the bridge for bidirectional adaptation.}
\label{fig:forward-backward-loop}
\end{figure}
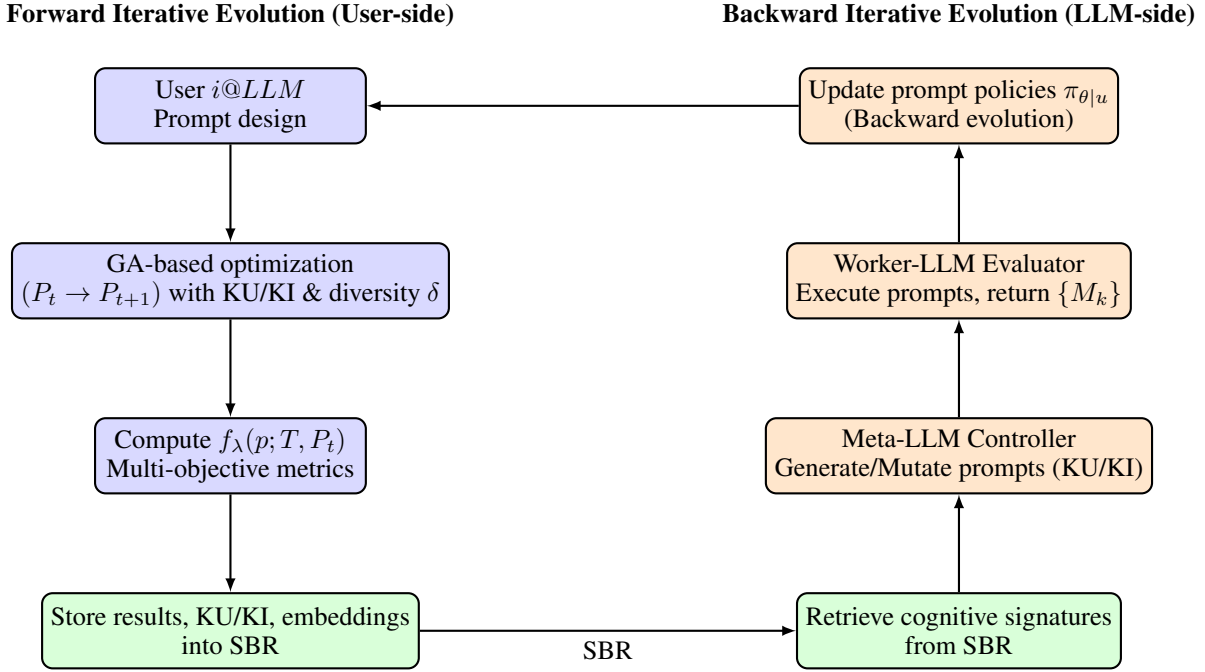

In this subsection, we describe the Forward and Backward iterative evolution processes for Subclass Brain formation using a GA-based approach. The coupling of forward and backward flows creates a \textit{closed-loop evolutionary process} in which Subclass Brains emerge organically. Section~\ref{sec:experiments} will present real-world experiments to illustrate these processes. Figure \ref{fig:forward-backward-loop}
shows the closed-loop Forward/Backward Iterative Evolution for Subclass Brain formation. The Subclass Brain Registry
(SBR) serves as the bridge for bidirectional adaptation.

\subsubsection{Forward Iterative Evolution (User-side)}
\label{subsec:forward}

Following the methodology in the Urban Air Mobility (UAM) Corridor in Brasilia (Eixão-UAM) study~\citep{weigang2025eixao}, we apply a hybrid \textbf{LLM+GA} framework to generate multi-objective decision solutions, thereby strengthening the user’s ability to design, evaluate and refine \textit{(prompt, fitness, solution)} triplets. This \emph{Forward Iterative Evolution} focuses on enhancing the user side of the dyad $User_i @ \mathrm{LLM}$, forming the basis for subsequent backward adaptation on the LLM side.

\paragraph{Air traffic control context.}
In Unmanned Aerial Vehicle (UAV) vertiport scheduling, capacity limitations are modeled by specifying the number of available takeoff pads per UAV class \citep{ferreira2014genetic}. Two contrasting scheduling algorithms are evaluated:
\begin{enumerate}
    \item \textbf{Round-Robin (RR):} A simple fairness-oriented online scheduler \citep{halder2022dynamic}.
    \item \textbf{Genetic Algorithm (GA):} A multi-objective optimizer balancing wait time, fairness and pad utilization.
\end{enumerate}

\paragraph{Round-Robin scheduling.}
The RR algorithm cycles through takeoff requests, allocating a fixed quantum $q=30\,\mathrm{s}$ per request:
\begin{equation}
\mathrm{Service}(P_i) = \min\big(q,\ t_i^{(c)}\big),
\end{equation}
\begin{equation}
t_i^{(c+1)} = t_i^{(c)} - q,
\end{equation}
where $t_i^{(c)}$ is the remaining service time in cycle $c$. RR ensures fairness and prevents monopolization.

\paragraph{GA-based optimization.}
Let $Q = \{a_1,\dots,a_{N_{\text{queue}}}\}$ denote the UAV queue. The GA searches over permutations $c \in S_Q$ to minimize \citep{alolaiwy2023multi}:
\begin{equation}
c^\star = \arg\min_{c\in S_Q} f_\lambda(c; T, P_t),
\end{equation}
where $f_\lambda$ is the multi-objective fitness function aligned with Section~\ref{subsec:formal-defs}. The cost terms include:
\begin{itemize}
    \item Average wait time $\bar{T}_{\text{wait}}$;
    \item Wait time standard deviation $\sigma_T$;
    \item $95$th percentile delay $T_{95\%}$;
    \item Time-varying pad penalties $w_k(t)\,T_{\text{penalty}}$.
\end{itemize}
Weights $\alpha_1,\alpha_2,\alpha_3$ are interactively tuned with the LLM to explore Pareto-optimal trade-offs between responsiveness, fairness and resource constraints.

\paragraph{LLM-assisted cost function evolution.}
Initial experiments showed that RR outperformed GA~v1 due to over-penalization of scarce pad usage. The LLM participated in:
\begin{enumerate}
    \item Diagnosing failure modes via log interpretation;
    \item Prototyping improved $f_\lambda$ variants (GA~v2--v5);
    \item Framing theoretical insights into online scheduling.
\end{enumerate}

The $f_\lambda$ evolution included:
\begin{itemize}
    \item \textbf{GA v1:} Baseline (wait time + fixed pad penalties);
    \item \textbf{GA v2:} Reduced Class~1 penalties;
    \item \textbf{GA v3:} Time-dependent $w_k(t)$;
    \item \textbf{GA v4:} Fairness-oriented (max delay penalty);
    \item \textbf{GA v5:} Statistical optimization (mean, std, $95$th percentile).
\end{itemize}

\paragraph{Generation of \texorpdfstring{$(\text{prompt}, f_\lambda, \text{solution})$}{(prompt, f, solution)} triplets.}
Each scenario is generated from a distinct prompt $p\in\mathcal{P}$ and evaluated with $f_\lambda$. The GA produces candidate sequences $c$, from which performance metrics $\{M_k\}$ are recorded. This yields a dataset:
\begin{equation}
\label{eq:equation 6}
\mathcal{R} = \{(p_i, f_\lambda(p_i;T,P_t), \text{solution}_i)\}_{i=1}^N,
\end{equation}
which is stored in the Subclass Brain Registry (SBR) for use in Backward Iterative Evolution (Section~\ref{sec:backward}).

\subsubsection{Backward Iterative Evolution (LLM-side)}
\label{sec:backward}

In the \emph{Backward Iterative Evolution}, the datasets $\mathcal{R}$ generated in the Forward phase are used to tune the LLM's response strategies via prompt optimization, enabling the model to internalize user-specific heuristics and cognitive patterns. This process represents the LLM-side adaptation in the dyad $(u @ \mathrm{LLM})$, complementing the user-side evolution described in Section~\ref{sec:iterative_evolution}.

\paragraph{Context.}
In Subclass Brain formation, prompt--response dynamics often unfold through trial-and-error adjustment of instructions, parameters and task structures. To accelerate and structure this process, we implement an \textbf{LLM-assisted GA framework} in which the LLM plays both \emph{controller} and \emph{worker}, evolving high-performing prompts under explicit diversity and explainability constraints.

\paragraph{Architecture.}
The framework comprises four key components:
\begin{enumerate}
    \item \textbf{Meta-LLM Controller}: Generates and mutates candidate prompts $p\in\mathcal{P}$ using performance feedback and KU/KI-guided constraints.
    \item \textbf{Worker-LLM Evaluator}: Executes $p$ on a task $T$ and returns evaluation metrics $\{M_k\}$.
    \item \textbf{Vector Embedding Bank}: Stores $\varphi(p)$ for diversity filtering, implemented via FAISS/Milvus.
    \item \textbf{KU/KI Filtering Mechanism}: Maintains two keyword sets:
    \begin{itemize}
        \item \textbf{KU (Key-Useful)}: Correlated with high $f_\lambda$ scores;
        \item \textbf{KI (Key-Irrelevant)}: Correlated with low $f_\lambda$ scores.
    \end{itemize}
\end{enumerate}

\paragraph{Evolutionary workflow.}
Let $P_t = \{p_1,\dots,p_n\}$ be the prompt population at generation $t$. Each $p_i$ is scored via the Worker-LLM:
\[
f_\lambda(p_i; T, P_t) \in \mathbb{R}.
\]
The next generation is:
\begin{equation}
P_{t+1} = \mathsf{Select}\!\left(\mathsf{Mutate}\big(\mathsf{Crossover}(P_t);\ \mathrm{KU},\mathrm{KI}\big)\right),
\end{equation}
where:
\begin{itemize}
    \item \texttt{Crossover}: recombines segments of top-ranked prompts;
    \item \texttt{Mutate}: applies lexical/structural changes under KU/KI constraints;
    \item \texttt{Select}: retains candidates with top $f_\lambda$ scores subject to diversity.
\end{itemize}
KU/KI lists are updated from top/bottom quartiles of $P_t$.

\paragraph{Diversity control.}
A candidate $p'$ is admissible only if:
\begin{equation}
\max_{q \in P_{t+1}} \mathrm{sim}\big(\varphi(p'), \varphi(q)\big) < \delta,
\end{equation}
where $\delta\in(0,1)$ is the diversity margin.

\paragraph{Optimization objectives.}
We jointly optimize:
\begin{itemize}
    \item \textbf{Task performance}: maximize $\{M_k\}$ (accuracy, F1, etc.);
    \item \textbf{Instruction economy}: minimize token cost $C_{\mathrm{tok}}(p)$;
    \item \textbf{Interpretability}: maintain traceable KU/KI--metric links.
\end{itemize}
Formally:
\begin{equation}
\max_{p \in \mathcal{P}} f_\lambda(p;T,P_t)
\quad \text{s.t. diversity \& explainability constraints.}
\end{equation}

\paragraph{Theoretical implication.}
This backward process simulates a \emph{non-human} Subclass Brain evolving under autonomous control, complementing user-guided Forward evolution. Patterns emerging from high-performing prompts and their KU/KI features reveal latent structures and biases in the LLM's internal knowledge. Comparing these across users or instances supports modeling of \emph{population-level dynamics} for Subclass Brain emergence, providing input to Swarm Intelligence (Section~\ref{sec:swarm}) and Superclass Brain formation (Section~\ref{sec:superclass_brain}).

\subsection{Swarm Intelligence}
\label{sec:swarm}

Swarm Intelligence (SI) is a branch of computational and artificial intelligence inspired by the collective behaviors of decentralized, self-organizing systems~\citep{kennedy2006swarm,nedjah2006swarm}. It models how simple agents, through local interactions, can give rise to complex global behaviors. Representative bio-inspired paradigms include:  
(1) \textbf{Ant Colony Optimization (ACO)} -- simulating pheromone-based path reinforcement during foraging;  
(2) \textbf{Particle Swarm Optimization (PSO)} -- emulating flocking dynamics to search high-dimensional spaces; and  
(3) \textbf{Bee Algorithms} -- inspired by cooperative foraging and resource allocation in honey bee colonies.  

In multi-agent systems, these principles are applied to decompose complex systems into smaller, interconnected components. Agents communicate, coordinate and negotiate to achieve shared objectives, enabling applications in UAV coordination, sensor networks, cooperative robotics, traffic flow optimization and dynamic resource scheduling~\citep{nogueira2014using,devi2024review}.

\vspace{0.5em}
\noindent\textbf{Swarm Intelligence for Subclass Brain Evolution.}  
In our LLM-based framework, SI emerges from the interaction of multiple \emph{Subclass Brains}--persistent cognitive pairs formed by high-quality users and an LLM (Section~\ref{sec:iterative_evolution}). Each Subclass Brain independently executes a Forward/Backward evolutionary loop (Figure~\ref{fig:forward-backward-loop}), producing optimized \texttt{(prompt, fitness, solution)} triplets. These outputs feed into a \emph{Swarm Layer} where candidate prompts, strategies or parameter settings are pooled, compared and evolved.

Inspired by the GA-based UAV scheduling in the Eixão-UAM project~\citep{weigang2025eixao}, we employ a \textbf{multi-objective fitness function} to evaluate swarm-level performance:
\begin{equation}
\mathrm{Fitness}_i = \sum_{j=1}^{n} w_j \cdot M_{i,j}
\end{equation}
where $M_{i,j}$ is the performance of the $i$-th Subclass Brain on the $j$-th metric and $w_j$ is the corresponding metric weight, reflecting domain-specific priorities.

Typical metrics include:
\begin{itemize}
    \item \textbf{Accuracy}: Task correctness and success rate.
    \item \textbf{Creativity}: Novelty and diversity of generated solutions.
    \item \textbf{Generality}: Cross-domain adaptability.
    \item \textbf{Robustness}: Stability under noise, perturbations or incomplete data.
\end{itemize}

The GA in the Swarm Layer applies \emph{selection}, \emph{crossover} and \emph{mutation} to the prompt/strategy pool, with a KU/KI-guided diversity control mechanism to prevent premature convergence. The resulting high-fitness artifacts are registered in the Subclass Brain Registry (SBR) for reuse and cross-agent distillation.

\begin{figure*}[h]
    \centering
    \includegraphics[width=0.65\textwidth]{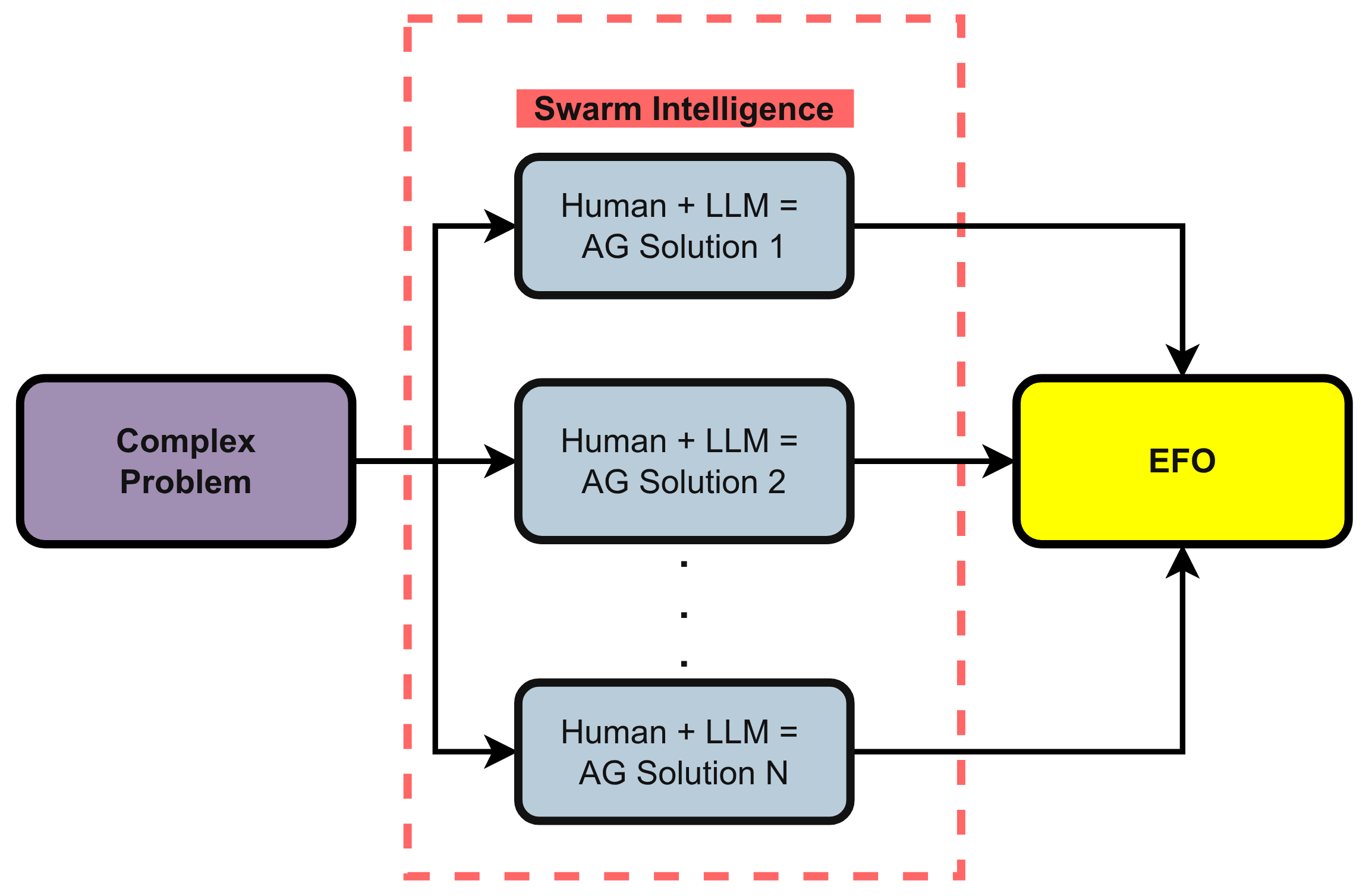}
    \caption{Swarm intelligence framework.}
    \label{fig:EFO}
\end{figure*}

\begin{table}[h]
\centering
\caption{Integrated architecture of the swarm intelligence framework}
\label{tab:swarm-architecture}
\begin{tabular}{lll}
\toprule
\textbf{Layer} & \textbf{Modules} & \textbf{Source} \\
\midrule
Swarm Layer & Prompt Pool + Evaluation & Student EFO design \\
Fitness Layer & Multi-objective Cost + Dashboard & Eixão-UAM GA framework \\
Brain Registry & Cognitive Signatures + KU/KI & Superclass Brain model \\
Meta-LLM Layer & Global Pattern Distillation & Integrated evolution \\
\bottomrule
\end{tabular}
\end{table}

Figure \ref{fig:EFO} shows the evolutionary framework optimization (EFO). A complex problem (purple) is addressed by multiple human--LLM cognitive pairs (blue), each producing candidate solutions. Within the Swarm Intelligence layer (red dashed border), these candidates are evolved using a GA module (yellow), applying multi-objective selection, crossover and mutation. The best evolved solution is fed back to the original problem, forming a closed-loop refinement process. The table \ref{tab:swarm-architecture} also shows the integrated architecture with four layers of the swarm intelligence framework.

\subsection{Superclass Brain}
\label{sec:superclass_brain}

\subsubsection{Main Steps Toward Superclass Brain Formation}

The \textit{Superclass Brain} is a higher-order collective intelligence structure emerging from the coordinated evolution of multiple \textit{Subclass Brains}.  
Rather than merging model parameters, it synthesizes \emph{cognitive patterns}, optimized \emph{prompt strategies} and \emph{domain-specific heuristics} through structured cross-agent interaction.  
The transition from isolated Subclass Brains to a unified Superclass Brain proceeds through four interdependent stages, each forming a closed-loop with preceding layers (Sections~\ref{sec:iterative_evolution}–\ref{sec:swarm}).

\paragraph{Step 1 – Standardized Cognitive Signatures.}
Each Subclass Brain, operating locally (e.g., via Ollama) or in a user-specific cloud instance, periodically exports an anonymized \emph{cognitive signature} capturing:
\begin{itemize}
    \item Curated prompt--response pairs with performance metadata;
    \item Representative success cases and quantitative task scores;
    \item Keyword-based cognitive features: \textbf{KU} (Key-Useful) and \textbf{KI} (Key-Irrelevant);
    \item Behavioral descriptors (e.g., “conservative”, “aggressive”, “exploratory”).
\end{itemize}

\noindent Example cognitive signature:
\begin{verbatim}
{
  "user": "User_i@GPT",
  "domain": "Energy Scheduling",
  "KU": ["multi-hop", "weather forecast", "delay tolerance"],
  "KI": ["verbose explanation", "poetic summary"],
  "Top5_Prompts": [
    {"prompt": "...", "response": "...", "score": 0.92},
    ...
  ]
}
\end{verbatim}

\paragraph{Step 2 – Subclass Brain Registry (SBR).}
A centralized or federated registry stores and indexes these signatures to:
\begin{itemize}
    \item Enable embedding-based semantic search and retrieval;
    \item Facilitate behavioral alignment and cross-domain prompt reuse;
    \item Aggregate high-quality datasets for collective fine-tuning.
\end{itemize}
The SBR acts as the shared \emph{cognitive memory} of the ecosystem.

\paragraph{Step 3 – Swarm Alignment Layer.}
Following Talebirad et~al.~\cite{talebiradwisdom}, multiple Subclass Brains execute the same task in parallel.  
Their outputs are reconciled via:
\begin{itemize}
    \item Multi-agent consensus mechanisms (voting, ranking, averaging);
    \item Cross-prompt fusion and knowledge distillation;
    \item Meta-agent coaching with reinforcement learning.
\end{itemize}
This layer can operate in both cloud-based orchestration platforms and decentralized, privacy-preserving federated setups.

\paragraph{Step 4 – Evolutionary Feedback to LLMs.}
The distilled swarm-level insights are transformed into:
\begin{itemize}
    \item High-performance prompt blueprints and style guides;
    \item Cross-user fine-tuning datasets for domain adaptation;
    \item Structured reinforcement templates for continual learning.
\end{itemize}
These artifacts may be fed back to foundation model providers (e.g., OpenAI, Anthropic) via RLHF or RLAIF pipelines, closing the loop between \emph{collective cognitive evolution} and \emph{model architecture refinement}.

\subsubsection{Formal Definitions of Subclass Brain, Superclass Brain and SuperBrain}
\label{subsec:formal-defs}

\paragraph{Tasks, prompts and evaluation.}
Let $T$ denote a task with input--output specification and let $\mathcal{P}$ be the prompt space. A \textit{Worker-LLM} executes a prompt $p\in\mathcal{P}$ on $T$ and returns $y=\mathrm{Worker\text{-}LLM}(p,T)$. For a set of evaluation metrics $\{M_k\}_{k=1}^K$ with weights $\mathbf{w}\in\mathbb{R}^K_+$, $\sum_{k=1}^K w_k=1$, the multi-objective fitness is:
\begin{equation}
f_\lambda(p;T,P_t)\;=\;\sum_{k=1}^K w_k\, M_k\!\big(\mathrm{Worker\text{-}LLM}(p,T)\big)\;-\;\lambda_{\mathrm{tok}}\,C_{\mathrm{tok}}(p)\;-\;\lambda_{\mathrm{div}}\,\Psi_\delta(p;P_t)\;-\;\lambda_{\mathrm{exp}}\,\Xi(p),
\end{equation}
where $C_{\mathrm{tok}}$ is token cost, $\Psi_\delta$ is a soft diversity penalty and $\Xi$ is an explainability regularizer (e.g., KU/KI traceability).

\paragraph{Subclass Brain.}
A \textit{Subclass Brain} associated with a high-value user $u$ is the tuple
\begin{equation}
\mathrm{SB}_u := \big(u,\ \mathcal{H}_u,\ \mathcal{M}_u,\ \pi_{\theta\mid u}\big),
\end{equation}
where $\mathcal{H}_u$ stores the interaction history, $\mathcal{M}_u$ is persistent memory and $\pi_{\theta\mid u}$ is the LLM’s response policy conditioned on $u$. Its \emph{cognitive signature} is:
\begin{equation}
\mathbf{c}_u := g\big(\mathcal{H}_u,\mathcal{M}_u\big)\in\mathbb{R}^d,
\end{equation}
including semantic centroids of top prompts/responses, KU/KI keyword statistics, behavioral tags, reliability $\rho_u\in[0,1]$ and token cost traces. The \textit{Subclass Brain Registry} (SBR) stores $\{\mathbf{c}_u\}_{u\in U}$.

\paragraph{Forward and backward evolution.}
\textit{Forward (user-side):} A genetic algorithm evolves prompt populations $P_t$ under KU/KI constraints and diversity margin $\delta$, producing $(p,y,f_\lambda)$ triplets and updating $\mathbf{c}_u$.  
\textit{Backward (LLM-side):} The Meta-LLM distills swarm patterns into a library $\Pi$ and updates $\pi_{\theta\mid u}$ or model-side parameters when allowed. Compactly:
\begin{equation}
\mathrm{SBR}_{t+1}=\mathrm{SBR}_t\cup\mathcal{R}(P_{t+1}),\quad
\Theta_{t+1}=\mathcal{U}(\Theta_t,\Pi_{t+1}).
\end{equation}

\paragraph{Superclass Brain.}
A \textit{Superclass Brain} is the aggregation of multiple Subclass Brains via a swarm alignment operator $\mathcal{A}$:
\begin{equation}
Q(p) \ \propto\ \sum_{u\in U}\alpha_u\, S\big(p,\mathbf{c}_u\big), \quad \alpha_u\propto \rho_u,
\end{equation}
where $S$ scores prompts against cognitive signatures and $Q\in\Delta(\mathcal{P})$ is a distribution over prompts. The Meta-LLM consumes $(\mathrm{SBR},Q)$ to produce a distilled pattern library $\Pi$, supporting collective reasoning that exceeds any individual $\mathrm{SB}_u$.

\paragraph{SuperBrain.}
The \textit{SuperBrain} is the coupled system:
\begin{equation}
\mathrm{SuperBrain} := \big( \{\mathrm{SB}_u\}_{u\in U},\ \mathcal{A},\ \mathcal{D},\ \mathcal{U} \big),
\end{equation}
where $\mathcal{A}$ is swarm aggregation, $\mathcal{D}$ is distillation into $\Pi$ and $\mathcal{U}$ is the update of policies and knowledge bases. Through repeated forward (user-side) and backward (LLM-side) evolution across the population, the system self-organizes into a distributed, adaptive cognitive entity whose problem-solving capacity surpasses both isolated LLM instances and any single Subclass Brain.

\section{Experiment of Forward Iterative Evolution (User-side)}
\label{sec:experiments}

As outlined in Subsection~\ref{subsec:forward}, the Forward Iterative Evolution process applies to UAV take-off sequence scheduling, where the user (Subclass Brain side) iteratively refines Genetic Algorithm (GA) configurations with LLM assistance.

\paragraph{Round Robin (RR) as a theoretical benchmark.}
In simulations, RR scheduling yielded the shortest average wait time and lowest worst-case delay, reflecting its fairness and stateless online design. However, real-world constraints--mission priorities, unpredictable weather, platform capacity limits and regulatory factors--make strict RR enforcement impractical. Thus, RR serves mainly as an upper-bound benchmark rather than a deployable solution.

\paragraph{Challenges of GA-based optimization.}
GAs are well-suited to multi-objective, multi-constraint scheduling, but practical performance hinges on:  
(1) generating a representative initial population,  
(2) defining a fitness function aligned with operational goals and  
(3) tuning parameters for variable conditions (e.g., wind speed, mission class).  
These design bottlenecks limit human-only approaches in dynamic environments.

\paragraph{LLM-assisted strategy generation.}
LLMs (e.g., GPT-4.5, GPT-4o mini, Gemini 2.5 Pro) enhance GA development by:  
(1) integrating local knowledge bases (RAG) such as Brasília traffic rules and weather history,  
(2) adapting to dynamic inputs (``Golden Quarter'' updates),  
(3) rapidly generating GA variants (v$_i$) through unified prompt templates and  
(4) proposing interpretable fitness function alternatives.  
In this setting, the LLM acts as a cognitive collaborator--hypothesis generator, experimental designer and rapid feedback provider--rather than a passive coding assistant.

\subsection{Experimental Background}
\label{sec:experimental_background}

To verify the feasibility and applicability of the proposed LLM+GA collaborative optimization approach, we first developed unified RR and GA versions of the UAV take-off sequence scheduling program, tested under the same Vertiport scenario. All experiments were conducted between late July and early August 2025 and were organized into three groups according to their research objectives:

\paragraph{(1) Prompt Variation on the Same LLM Platform.}
On the same LLM platform (e.g., Gemini~2.5~Pro), we modified the sorting optimization objectives and constraints in the Prompt instructions, thereby adjusting the GA parameters and producing different take-off sequence results. As described in Section~3.2.1, GAv1--v5 correspond to five different parameter combinations generated on the same platform. This group aims to analyze how varying Prompt conditions affect optimization results and to compare the relative performance of the resulting solutions.

\paragraph{(2) Cross-Platform Comparison of Results.}
Keeping the GA scheduling program and Prompt instructions exactly the same, we tested different LLM platforms (GPT-4.0, GPT-4.o~mini, Grok~3, GPT-5.0 and Claude~4) to generate GA parameters (GA~v6--v11) and observed the resulting differences in scheduling performance. The goal is to assess the stability and sensitivity of the LLM+GA method across platforms. If the results are broadly similar, it indirectly supports the method’s universality and feasibility.

\paragraph{(3) Impact of User Expertise and Interaction Depth.}
Using the same GA program and Prompt instructions, different users---with varying backgrounds, expertise and depth of interaction with LLMs---ran the experiments to generate GA parameters (GA~v6--v11). The aim is to investigate how user qualifications and interaction patterns affect optimization outcomes. The participants were:
\begin{itemize}
    \item \textbf{LRS} (senior undergraduate in computer science): LRS@Gemini~2.5~Pro (GA~v1--v5), LRS@GPT-5.0 (GA~v9), LRS@Claude~4 (GA~v11);
    \item \textbf{LWG} (senior AI and ATC/ATM expert): LWG@GPT-4.0 (GA~v7), LWG@Grok~3 (GA~v8);
    \item \textbf{JRS} (high school student): JRS@GPT-5.0 (GA~v10).
\end{itemize}

This experiment set is particularly designed to compare the performance differences between ordinary users and ``valuable users'' (highly experienced and frequent, in-depth LLM collaborators) in co-developing optimized scheduling strategies.

On the other hand, GA v7--v11 were all generated under exactly the same prompt set, ensuring that the LLM platform and user expertise were the only changing factors.  
The original prompts, written by the student, are preserved here without correction to reflect the authentic experimental conditions:

\begin{itemize}
    \item \textbf{Prompt 1:} Analyze the following code: (whole GA for take-off sequence scheduling code).
    \item \textbf{Prompt 2:} I want you to modify this code in order for it to run your suggestions in the same way it has run Gemini's, that is, I want you to write your suggestions in such a way that it can be tested side by side with the other one's, in an independent way and sucessfully analyzed.
    \item \textbf{Prompt 3:} I want a balanced approach, but minimizing average wait time should be a priority.
\end{itemize}

This strict prompt consistency allows the subsequent analysis (Sections~\ref{subsec:interative}--\ref{subsec:comparison}) to isolate the effects of LLM platform variation and user expertise, ensuring that observed differences in GA performance can be attributed primarily to these factors rather than prompt design.

\subsection{GA v1-5 results from a scientific dialogue}
\label{subsec:interative}

The GA refinement followed a three-phase human–AI workflow \citep{weigang2025eixao}:

\textbf{Phase 1: Diagnosis.}  
Baseline GA performance lagged behind RR in both average and worst-case metrics. Numerical outputs and plots were presented to the LLM, which identified errors (e.g., mislabelled curves) and self-corrected upon clarification.

\textbf{Phase 2: Hypothesis generation.}  
Guided by the researcher, the LLM proposed new fitness functions incorporating fairness, statistical robustness and time-weighted penalties. Five GA variants (v1–v5) were generated and tested in a single session. Table \ref{tab:GA_v1-5_results} shows the take-off sequence scheduling results for GA variants v1–v5, with improvement rates relative to GA v1 \citep{weigang2025eixao}.

\textbf{Phase 3: Insight extraction.}  
The LLM assisted in interpreting trade-offs (e.g., GA v5’s 60\% reduction in worst-case delay over v1) and relating them to design decisions, while discarding ineffective variants (e.g., v2, v3).

\begin{table*}[htbp]
    \centering
    \caption{Take-off sequence scheduling results for GA variants v1--v5, with improvement rates relative to GA v1}
    \label{tab:GA_v1-5_results}
    \footnotesize
    \resizebox{\linewidth}{!}{
        \begin{tabular}{lcccccc}
            \hline
            \textbf{Metric} & \textbf{RR} & \textbf{GA v1} & \textbf{GA v2} & \textbf{GA v3} & \textbf{GA v4} & \textbf{GA v5} \\
            \hline
            Avg. Wait Time       & \textbf{00m 03s} & 00m 10s & 00m 10s & 00m 10s & 00m 09s & 00m 09s \\
            Max. Wait Time       & \textbf{03m 26s} & 12m 18s & 17m 28s & 18m 11s & 06m 33s & 04m 58s \\
            \% No Wait           & \textbf{77.39\%} & 60.00\% & 60.00\% & 60.00\% & 59.99\% & 59.94\% \\
            \% Long Wait ($>$2min) & \textbf{0.06\%} & 0.31\% & 0.30\% & 0.30\% & 0.27\% & 0.34\% \\
            \hline
            Improvement in Avg. Wait Time (\%) & -- & 0.00\% & 0.00\% & 0.00\% & 10.00\% & 10.00\% \\
            Improvement in Max. Wait Time (\%) & -- & 0.00\% & -42.02\% & -47.45\% & 46.93\% & 59.82\% \\
            Improvement in \% No Wait          & -- & 0.00\% & 0.00\% & 0.00\% & -0.02\% & -0.10\% \\
            Improvement in \% Long Wait        & -- & 0.00\% & 3.23\% & 3.23\% & 12.90\% & -9.68\% \\
            \hline
        \end{tabular}
    }
\end{table*}

\subsection{Comparison Analysis between GA v1-v5 and GA v6--v11 Results }\label{subsec:comparison}

Tables~\ref{tab:GA_v1-5_results} and~\ref{tab:ga_v6_v11} present the performance of GA variants v1-5 and v6--v11 in the UAV take-off sequence scheduling task, measured by four key metrics: \textit{Average Wait Time}, \textit{Maximum Wait Time}, \textit{\% No Wait} and \textit{\% Long Wait ($>$2 min)}. These results confirm the practical engineering applicability of the proposed LLM+GA evolutionary approach: by modifying prompt instructions and tuning fitness function parameters, the GA can approach near-optimal scheduling outcomes. For instance, GA v5 (Gemini 2.5 Pro) achieved a Max. Wait Time of 4m58s, while GA v8 (Grok 3.0) reached 4m56s, both approaching the ideal RR benchmark of 3m26s. This demonstrates that the LLM+GA iterative process can progressively converge toward optimal take-off sequence scheduling across different LLM platforms.

From another perspective, this experiment introduces a “human cognitive intervention” variable by controlling the LLM model version and prompt while varying the expertise level and interaction depth of human collaborators. Even with identical models and prompts, user expertise significantly influenced GA performance:
\begin{itemize}
    \item \textbf{Experts (LWG)} incorporated fine-grained operational constraints (e.g., ATC rules, wind speed thresholds, vertiport distribution), directly impacting initial population design and fitness function tuning. This is evident in GA v7 (GPT-4.0) and GA v8 (Grok 3.0), both achieving Avg. Wait Time of 8s and No Wait rates exceeding 61.8\%. Notably, GA v8 reached the best Max. Wait Time (4m56s), a 60\% improvement over GA v1.
    \item \textbf{Undergraduate student (LRS)} emphasized rapid implementation and general-purpose strategies, paying less attention to complex constraints. Nevertheless, his iterative work with Gemini 2.5 Pro (GA v1--v5) produced competitive results, with GA v5 achieving a Max. Wait Time of 4m58s.
    \item \textbf{Secondary school student (JRS)} used more intuitive and less formal constraints, leading the LLM to different optimization pathways. GA v10 underperformed across most metrics, reflecting a weaker constraint formulation and reduced parameter tuning.
\end{itemize}

%Figures~X1 and~X2 illustrate the performance distributions: in GA v7 and GA v8, the Avg. Wait Time for Class-1 (large UAVs) approaches RR’s ideal values and their hour-by-hour Avg. Wait Times remain close to RR throughout most of the day.

\begin{table*}[htbp]
    \centering
    \caption{Take-off sequence scheduling results for GA variants v6--v11, with improvement rates relative to GA v1}
    \label{tab:ga_v6_v11}
    \footnotesize
    \resizebox{\linewidth}{!}{
        \begin{tabular}{lcccccc}
            \hline
            \textbf{Metric} & \textbf{GA v6} & \textbf{GA v7} & \textbf{GA v8} & \textbf{GA v9} & \textbf{GA v10} & \textbf{GA v11} \\
            \hline
            Avg. Wait Time       & 00m 10s & 00m 08s & 00m 08s & 00m 12s & 00m 12s & 00m 12s \\
            Max. Wait Time       & 10m 36s & 08m 51s & \textbf{04m 56s} & 10m 32s & 12m 11s & 10m 06s \\
            \% No Wait           & 60.24\% & 61.87\% & 61.82\% & 60.09\% & 60.11\% & 60.06\% \\
            \% Long Wait ($>$2min) & 0.35\% & 0.16\% & \textbf{0.08\%} & 1.24\% & 1.16\% & 1.48\% \\
            \hline
            \textbf{Improvement in Avg. Wait Time (\%)} & 0.0\% & +20.0\% & +20.0\% & -20.0\% & -20.0\% & -20.0\% \\
            \textbf{Improvement in Max. Wait Time (\%)} & +14.1\% & +28.2\% & +59.9\% & +14.8\% & +0.95\% & +18.0\% \\
            \textbf{Improvement in \% No Wait (\%)}     & +0.40\% & +3.12\% & +3.03\% & +0.15\% & +0.18\% & +0.10\% \\
            \textbf{Improvement in \% Long Wait (\%)}   & -12.9\% & +48.4\% & +74.2\% & -300.0\% & -274.2\% & -377.4\% \\
            \hline
        \end{tabular}
    }
\end{table*}

\textbf{Additional comparative insight is provided in Table~\ref{tab:GA_v1-5_results} and Table~\ref{tab:ga_v6_v11}, which report improvement rates over GA v1 for each metric.}  
The combined breakdown across GA v1--v11 highlights that:
\begin{itemize}
    \item Among GA v1--v5, GA v5 (Gemini 2.5 Pro) delivers the best overall performance, with a +59.8\% gain in \textit{Max. Wait Time} and a +12.9\% improvement in \textit{\% Long Wait}, approaching RR’s benchmark in worst-case delay.
    \item GA v4 also shows a substantial +46.9\% improvement in \textit{Max. Wait Time}, while maintaining Avg. Wait Time at +10\% better than GA v1.
    \item Among GA v6--v11, GA v8 (LWG@Grok 3) achieves the highest gains in \textit{Max. Wait Time} (+59.9\%) and \textit{\% Long Wait} (+74.2\%), representing the best performance across all variants v1--v11.
    \item GA v7 (LWG@GPT-4.0) presents a balanced profile, with +20\% improvement in Avg. Wait Time and +48.4\% in \textit{\% Long Wait}, second only to GA v8 in extreme delay reduction.
    \item GA v6 performs close to GA v1 in Avg. Wait Time but shows modest gains in \textit{Max. Wait Time}.
    \item GA v9 and GA v10 regress in certain metrics-particularly GA v10, which shows a large negative change in \textit{\% Long Wait} (-274.2\%), indicating significantly more long delays than GA v1.
    \item GA v11 moderately improves \textit{Max. Wait Time} but suffers from a similar regression in \textit{\% Long Wait} as GA v9 and GA v10.
\end{itemize}

These results reinforce three central conclusions:  
(i) LLM-assisted GA evolution benefits most from expert-driven parameter design and multi-round iterative refinement, as shown by GA v7 and GA v8;  
(ii) Aligning the fitness function with operational goals is more decisive than algorithmic complexity, as evidenced by GA v5’s performance despite using a simpler structural design;  
(iii) Although this study is based on a small sample, it spans multiple LLM platforms (four distinct models) and diverse user profiles (valuable and ordinary users) in a complete prompt–parameter evolution–take-off scheduling workflow, with consistent performance trends observed across conditions. This provides a solid foundation for larger-scale, multi-sample experiments in future work.  

According to equation~\ref{eq:equation 6}, these results yield the \texorpdfstring{$(\text{prompt}, f_\lambda, \text{solution})$}{(prompt, f, solution)} triplets, which serve as the input for the next stage--Backward Iterative Evolution on the LLM side--to further refine strategies through autonomous prompt optimization.

\section{Framework of Backward Iterative Evolution (LLM-side)}\label{sec:backward-iter-evo}

While the forward iterative evolution loop (Section~\ref{subsec:forward}) emphasizes human-driven exploration of prompts and task-specific heuristics, the \textbf{backward loop} formalizes how LLM-side mechanisms integrate user-derived feedback and swarm-level signals into the model update process. This section focuses on refining the cost-function weights within a scheduling framework, thereby illustrating how Subclass Brain heuristics can be aggregated and transferred toward a collective Superclass Brain. The experiment refines only the weights of the scheduling cost while keeping the inner optimization strictly deterministic and strongly convex. The cost of a candidate schedule $c$ follows the article's notation and semantics:
\begin{equation}
    f(c)=\alpha_{1}\,\overline{T}_{\text{wait}}(c)+\alpha_{2}\,\sigma_{T}(c)+\alpha_{3}\,T_{95\%}(c)+\sum_{k}w_k(t)\,T_{\text{penalty}}(c),\label{eq:6paper}
\end{equation}
where $\overline{T}_{\text{wait}}$ is the average waiting time (system responsiveness), $\sigma_{T}$ is the standard deviation of waiting times (fairness and dispersion), $T_{95\%}$ represents near worst-case delay (implemented through $\mathrm{CVaR}_{0.95}$ but kept in notation for continuity), $w_k(t)$ is a time-varying weight attached to pad $k$ at time $t$ (resource scarcity, priorities or congestion) and $T_{\text{penalty}}$ is the aggregated penalty due to contention or constraint violations. The coefficients $\alpha_1,\alpha_2,\alpha_3$ live on the simplex and express the operational trade-off among mean, dispersion and tail.

\begin{figure*}[h]
    \centering
    \includegraphics[width=\textwidth]{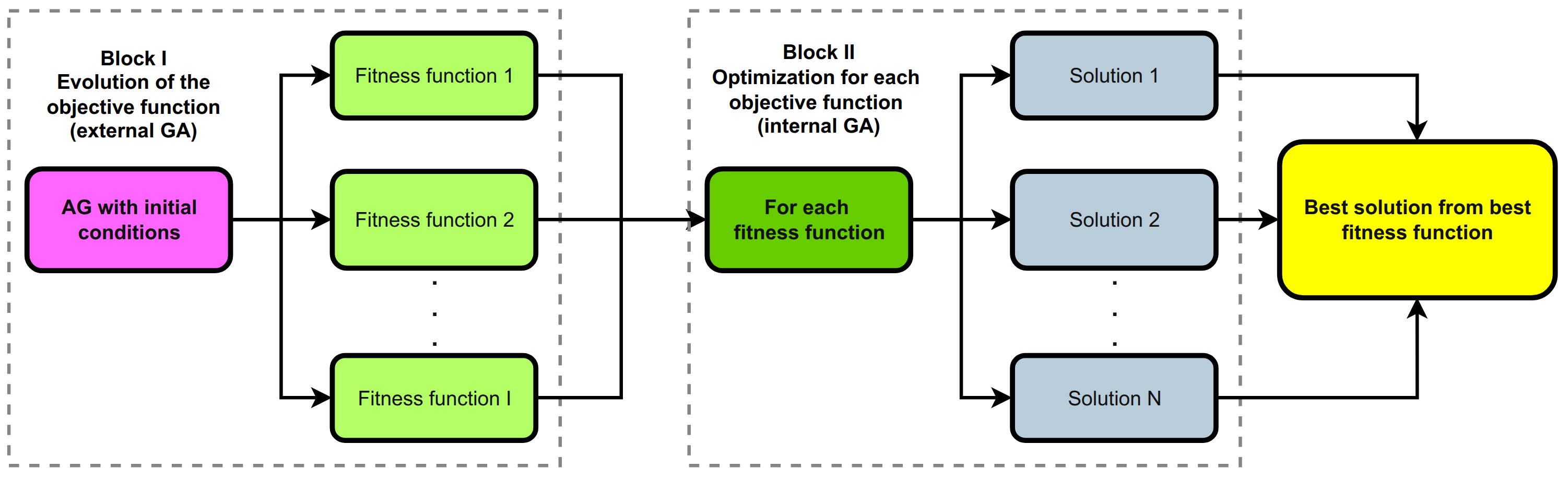}
    \caption{Bilevel GA with Dirichlet–Multinomial weight sampling and LSE-based convex interactions for strongly convex inner optimization.}
    \label{fig:efo}
\end{figure*}

As illustrated in Figure~\ref{fig:efo}, the proposed \textbf{Bilevel-GA} pipeline comprises two tightly coupled blocks. In \textbf{Block I}, an external genetic algorithm samples and evolves candidate cost functions $F(\cdot;\alpha)$ via a Dirichlet–Multinomial scheme, enabling convex interactions among $\overline{T}_{\text{wait}}$, $\sigma_T$, $T_{95\%}$ and $T_{\text{penalty}}$ through $\operatorname{LSE}_{\tau}$ aggregators, while enforcing $\alpha_{\{4\}}\ge\eta>0$ to guarantee strong convexity of the inner problem. In \textbf{Block II}, for each generated cost function, the convex scheduling program in the continuous variables $z$ (with capacity constraints) is solved by a second-order method to obtain the unique optimum $z^\star(\alpha)$. Performance indicators are then computed and the best candidate retained according to responsiveness as primary criterion and dispersion/tail penalties as tie-breakers. This setup allows the backward loop to integrate behavioral signals into cost-function refinements that persist beyond a single user or prompt.

\subsection{Block I: Strongly Convex Construction}\label{subsec:inner}

To solve scheduling with guarantees, we move from discrete sequences to a convex relaxation in continuous decision variables $z$ that induce the waiting-time vector $T(z)\in\mathbb{R}^n$ and per-pad utilizations $u_{k,t}(z)\in[0,u_{\max})$ under a convex feasible set $\mathcal{Z}$ of flow and capacity constraints. We recast the four primitives of \eqref{eq:6paper} as convex atoms
\begin{equation}
g_1(z)=\overline{T}_{\text{wait}}(z),\qquad
g_2(z)=\frac{1}{\sqrt{n}}\big\|P\,T(z)\big\|_2\ \text{with }P=I-\tfrac1n\mathbf{1}\mathbf{1}^{\!\top},
    \end{equation}
\begin{equation}
g_3(z)=T_{95\%}(z),\qquad
g_4(z)=\sum_{k,t} w_k(t)\,\phi\!\big(u_{k,t}(z)\big).   
\end{equation}

The mapping for $g_1$ is linear in $T(z)$. The map $g_2$ is a norm of a linear transform and remains convex. The tail proxy $g_3$ is implemented through $\mathrm{CVaR}_{0.95}$ on the observed waiting-time samples but reported as $T_{95\%}$ notation. The penalty $g_4$ uses a nondecreasing $m$-strongly convex function $\phi:[0,u_{\max})\to\mathbb{R}_+$ (for instance $\phi(u)=u^2/(1-u)$), which injects curvature through the utilizations $u_{k,t}(z)$. To model interactions among the primitives without leaving the convex regime, we aggregate any nonempty subset $S\subseteq\{1,2,3,4\}$ by the log-sum-exp operator
\begin{equation}
H_S(z)=\operatorname{LSE}_{\tau}\!\Big(\{g_i(z)\}_{i\in S}\Big)=\tau\log\!\sum_{i\in S}\exp\!\big(g_i(z)/\tau\big)\quad(\tau>0),
\end{equation}
a convex and monotone envelope that smoothly emphasizes the largest elements as $\tau\downarrow0$. The inner objective is the convex combination
\begin{equation}
F(z;\alpha)=\sum_{S\neq\varnothing}\alpha_S\,H_S(z),\qquad \alpha_S\ge0,\ \sum_{S}\alpha_S=1,\ \alpha_{\{4\}}\ge\eta>0,
\end{equation}
with $M=15$ weights (four singletons, six pairs, four triplets and one quadruplet). Because $H_{\{4\}}(z)=g_4(z)$ is $m$-strongly convex and appears with floor weight $\alpha_{\{4\}}\ge\eta$, the composite objective is $\mu$-strongly convex with $\mu\ge\eta\,m$ and therefore admits a unique global minimizer. The inner program reads
\begin{equation}
\min_{z\in\mathcal{Z}}\,F(z;\alpha)\ \ \text{subject to}\ \ 0\le u_{k,t}(z)<u_{\max},
\end{equation}
and is solved by a second-order or interior-point routine with backtracking line search.

\subsection{Block II: Dirichlet–Multinomial Weight Evolution}\label{subsec:outer}

The outer GA explores the of interaction weights $\alpha=(\alpha_S)_{S\neq\varnothing}$ while preserving convexity and curvature. Individuals carry $M$ genes that sum to one. For initialization or mutation around a parent vector $\alpha^{(\mathrm{par})}$, we draw $p\sim\mathrm{Dirichlet}(a)$ with $a_j=\kappa/M$ (symmetric start) or $a_j=\kappa\,\alpha^{(\mathrm{par})}_j$ (local exploration), then sample counts $n\sim\mathrm{Multinomial}(m_{\text{samp}},p)$ and set raw weights $\tilde\alpha_j=n_j/m_{\text{samp}}$. To guarantee $\alpha_{\{4\}}\ge\eta$ without rejection we apply the $\eta$-safe projection
\begin{equation}
\alpha=\eta\,e_{\{4\}}+(1-\eta)\,\tilde\alpha,
\end{equation}
which keeps nonnegativity and the unit sum. Each GA generation uses tournament selection (size $s\in\{2,3\}$), uniform arithmetic crossover on the (probability $p_c\in[0.7,0.9]$) followed by the same $\eta$-safe projection and Dirichlet–Multinomial mutation (probability $p_m\in[0.1,0.3]$) with concentration $\kappa\in[1,10]$ and sample size $m_{\text{samp}}\in\{16,32,64\}$. Population size $N_{\text{pop}}\in\{40,60,80\}$, generations $G\in[60,120]$ and elitism $N_{\text{elite}}\in\{1,2\}$ complete the controls. The fitness of an individual is the out-of-sample mean waiting time obtained by solving the inner program, namely $\text{fit}(\alpha)=\overline{T}_{\text{wait}}\big(z^\star(\alpha)\big)$, with ties broken by $\sigma_T$, $T_{95\%}$ and $T_{\text{penalty}}$ so that responsiveness remains the primary objective and fairness and tail risk are disciplined.

\subsection{Validation and Reporting}\label{subsec:validation}

Data are split into train, validation and test, all stratified by UAV class and weather condition to reflect heterogeneous regimes. The GA only observes validation scores when ranking individuals. Final reporting uses test and presents per-condition panels for $\overline{T}_{\text{wait}}$, $\sigma_T$, $T_{95\%}$ and throughput. The expected outcome is a consistent reduction in average waiting time together with controlled dispersion and tail across adverse scenarios; stability follows from strong convexity $\mu\ge\eta\,m$ and the curvature embedded in $g_4$ through the utilization penalty.

\subsubsection{Discussion and Relation to Backward Evolution}

This bilevel procedure illustrates how LLM-side mechanisms absorb Subclass Brain heuristics and project them into swarm-level refinements. In contrast to forward evolution, which relies on user-driven exploration, backward evolution systematically distills swarm-derived feedback into model-side optimization--closing the loop between local interaction and collective adaptation. Similar iterative refinement paradigms have been explored in \citet{ouyang2022training}, \citet{christiano2017deep}, \citet{shinn2023reflexion} and \citet{madaan2023selfrefine}, while the swarm-level integration resonates with \citet{burton2024large} and knowledge distillation methods \citep{hinton2015distilling,moslemi2024survey}. By embedding these mechanisms into the Superclass Brain registry, we ensure that individual user contributions are not lost but recursively shape the emergent meta-intelligence.

\section{Discussion and Architectural Implications}
\label{sec:discussion}

This section synthesizes the key implications of the proposed \textit{SuperBrain} paradigm from three complementary perspectives. 
First, Subsection~\ref{sec:Engineering} examines the \textbf{engineering significance} of the ``human creativity amplification + multi-strategy generation + autonomous evolutionary optimization + explainable results'' workflow, highlighting its potential to accelerate complex system design and optimization in real-world scenarios. 
Second, Subsection~\ref{sec:brain_comparison} provides a \textbf{comparative analysis} between \textit{SuperBrain} and two landmark large-scale AI paradigms--\textit{CAM-Brain} and \textit{AI 2027}--to position the proposed model within the broader trajectory of AI evolution. 
Finally, Subsection~\ref{sec:architectural_implications} explores \textbf{architectural recommendations} for future LLM/Transformer designs, emphasizing mechanisms such as long-term user-level memory, multi-agent internal interaction and meta-layer integration to support persistent cognitive evolution toward AGI-level intelligence.

\subsection{Engineering implications of the human–LLM co-evolution paradigm}
\label{sec:Engineering}

The proposed paradigm--\textit{human creativity amplification + multi-strategy generation + autonomous evolutionary optimization + explainable results}--demonstrates that engineering creativity can be systematically amplified through large language models (LLMs). By integrating the cognitive strengths of LLMs with the search and optimization capabilities of Genetic Algorithms (GAs), it offers a coherent framework for tackling complex system design and optimization challenges. In particular, it addresses two long-standing bottlenecks in GA applications: (i) initialization of a high-quality candidate population and (ii) design of domain-aligned fitness functions.

In the case study of Urban Air Mobility (UAM) scheduling \citep{weigang2025eixao}, this paradigm proved capable of generating, evaluating and refining multiple GA strategies under dynamic constraints, delivering measurable improvements in worst-case performance. The approach is generalizable to other domains requiring high-performance decision-making under uncertainty.

The key engineering implications are as follows:

\begin{enumerate}
    \item \textbf{Accelerating engineering iterations.}
    By automating the generation of initial conditions, fitness functions and parameter settings, the LLM--GA workflow enables rapid prototyping of diverse strategies. This shortens the design–test–refine cycle, which is critical in dynamic domains like UAM scheduling where external factors (e.g., weather, traffic patterns, regulatory constraints) change rapidly.

    \item \textbf{Amplifying human creativity.}
    The paradigm positions the LLM as a \textit{cognitive amplifier} rather than a passive code generator. Through contextual understanding and hypothesis generation, LLMs can propose novel optimization strategies and solution pathways that extend beyond the human designer’s initial search space.

    \item \textbf{Supporting complex system optimization.}
    While demonstrated in UAM scheduling, the approach scales to broader domains such as intelligent transportation, disaster response and energy management. These domains typically involve multi-objective optimization with dynamic, interdependent constraints--conditions under which the LLM+GA framework is particularly effective.

    \item \textbf{Enabling the evolution of collective intelligence.}
    Extending beyond single-task optimization, the paradigm can be embedded into multi-agent, multi-task ecosystems where multiple \textit{Subclass Brains} collaborate and co-evolve strategies. This opens pathways toward city-scale or even global-scale digital twin environments, redefining AI’s role from \textit{task executor} to \textit{strategic collaborator}.
\end{enumerate}

Overall, the human–LLM co-evolution paradigm offers immediate practical value in engineering workflows while also pointing toward a future where AI systems participate as adaptive, co-creative partners in collective intelligence architectures. Its broader significance becomes evident when compared with other large-scale AI paradigms, such as \textit{CAM-Brain} and \textit{AI 2027}, as discussed in Section~\ref{sec:brain_comparison}.

\subsection{Comparative Analysis: SuperBrain vs. CAM-Brain vs. AI 2027}
\label{sec:brain_comparison}

To contextualize the proposed \textit{SuperBrain} framework, we compare it against two influential paradigms in large-scale artificial intelligence development: the \textbf{CAM-Brain} project~\citep{buller1998brain}, a pioneering attempt at hardware-based neuroevolution and \textbf{AI 2027}~\citep{kokotajloAI2027}, a forward-looking vision of multi-agent collaborative intelligence for societal foresight. These three frameworks represent distinct approaches to scaling AI capabilities--hardware-centric autonomous evolution (CAM-Brain), cloud-based multi-agent orchestration (AI 2027) and human--LLM integrated cognitive evolution (SuperBrain). 

Table~\ref{tab:brain_comparison} summarizes their key characteristics across seven dimensions: primary goal, core mechanism, knowledge aggregation, user involvement, optimization strategy, scalability and innovation points.

\begin{table*}[htbp]
\centering
\caption{Comparison of SuperBrain, CAM-Brain and AI 2027 frameworks}
\label{tab:brain_comparison}
\begin{tabular}{p{3cm}p{4cm}p{4cm}p{4cm}}
\toprule
\textbf{Dimension} & \textbf{SuperBrain (This work)} & \textbf{CAM-Brain} \citep{buller1998brain} & \textbf{AI 2027} \citep{kokotajloAI2027} \\
\midrule
\textbf{Primary Goal} 
& Human–LLM integrated cognitive evolution; scalable collective intelligence via multi-layer architecture 
& Large-scale evolution of neural networks in hardware (FPGA) for autonomous agents 
& Predictive and collaborative AI ecosystem for societal, economic and scientific foresight \\

\textbf{Core Mechanism} 
& Forward \& backward iterative evolution of prompts/strategies; swarm intelligence layer; knowledge distillation into Superclass Brain 
& Cellular Automata Machine (CAM) hardware evolves neural architectures using genetic algorithms 
& Multi-agent wisdom extraction through structured prompt-response aggregation and consensus building \\

\textbf{Knowledge Aggregation} 
& Subclass Brain registry + KU/KI feature tracking; cross-user evolutionary feedback 
& Hardware-level synaptic evolution; direct fitness-based selection 
& Cross-domain data fusion and prompt alignment via meta-agents \\

\textbf{User Involvement} 
& Continuous human–AI co-creation; user-specific cognitive signatures 
& Minimal--focus on autonomous hardware learning 
& Human experts as domain validators and scenario designers \\

\textbf{Optimization Strategy} 
& Multi-objective GA with diversity control and interpretability constraints 
& Evolutionary search over network topology and weights 
& Iterative consensus and refinement through multi-agent simulation \\

\textbf{Scalability} 
& Distributed Subclass Brains with registry-based coordination; cloud/federated deployment 
& Limited by FPGA resources and simulation environment 
& Scales via cloud-based multi-agent orchestration \\

\textbf{Innovation Points vs. Others} 
& User-centered, explainable swarm evolution; bidirectional (forward/backward) learning loop 
& Hardware-oriented neuroevolution pioneer; fine-grained evolutionary control at circuit level 
& Socio-technical integration of AI for foresight and planning \\

\bottomrule
\end{tabular}
\end{table*}

As shown in Table~\ref{tab:brain_comparison}, \textit{SuperBrain} distinguishes itself from CAM-Brain and AI 2027 in three notable ways. 
First, it is explicitly \textbf{human-centered}: user-specific \textit{Subclass Brains} capture cognitive signatures, enabling personalized learning loops and explainable optimization strategies. 
Second, it adopts a \textbf{bidirectional evolutionary process}--combining forward user-driven iteration with backward LLM-side adaptation--whereas CAM-Brain focuses on hardware-level neuroevolution and AI 2027 emphasizes macro-scale agent consensus. 
Third, \textit{SuperBrain} incorporates \textbf{interpretability constraints} into its optimization pipeline, maintaining traceable links between prompt features (KU/KI) and performance metrics, which is absent in the other two frameworks.

This comparison underscores that \textit{SuperBrain} is neither a purely hardware-bound approach nor a purely cloud-based consensus system. Instead, it integrates human creativity, swarm intelligence and evolutionary optimization into a coherent architecture, positioning it as a bridge between micro-level cognitive augmentation and macro-scale collective intelligence.

\subsection{Architectural implications for future LLM/Transformer design}
\label{sec:architectural_implications}

While current LLMs have significantly advanced AI adoption and capabilities, their continued healthy evolution requires sustained architectural optimization and ecosystem development.  
In recent years, multiple extensions have emerged to enhance LLM adaptability and performance, such as \textit{Fine-Tuning}, \textit{RAG}, \textit{RAFT} \citep{zhangraft,di2024slim} and the \textit{Golden Quarter} \citep{bezerra2025llmquoter} methodology.  
Building upon these, the proposed \textit{SuperBrain} framework is not only valuable at the application level but will also have natural implications for the optimization of LLM’s underlying architecture--especially the Transformer \citep{vaswani2017attention}.  
The following suggestions outline potential architectural enhancements.

\subsubsection{Long-term, controllable user-level memory interface}

Most existing LLMs still have substantial limitations in storing and retrieving long-term user interaction histories.  
For example, GPT allows selective retention of certain conversation threads; Grok generally stores only the most recent 25 interactions.  
None of these systems currently maintain a systematic, controllable and persistent \textit{user-level cognitive memory}, nor do they integrate an \textit{RLHF/RLHB} module that learns from such histories to promote the emergence of a \textit{Superclass Brain}.

At the Transformer architectural level, we propose a \textbf{long-term, controllable user-level memory interface} with versioned local data import/export capabilities.  
This would enable valuable user-generated historical data to be stored, retrieved and processed under strict privacy safeguards, supporting sustained learning and personalized cognitive growth beyond short-term session memory.

\subsubsection{Multi-agent internal interaction modules--toward a ``wise'' LLM}

In the early internet era, the intrinsic value of data was underestimated.  
With the rise of \textit{big data}, the focus shifted toward data aggregation and monetization, yet flows remained largely one-way and lacked deep human–machine co-creation.

In the LLM era, large-scale pre-training leverages big data to create ``big models'' and enables more natural interaction.  
However, despite improvements through fine-tuning and RLHF, \textbf{post-training evolutionary adaptation}--continuous learning after deployment--remains limited.  
Users benefit from LLMs, but the reverse flow of new knowledge from users to LLMs is still minimal.

The \textbf{Forward/Backward Iterative Evolution for Subclass Brain} mechanism proposed in this work addresses this gap by embedding internal GA or strategy evolution modules and aggregating them via \textit{swarm intelligence} into a \textit{Superclass Brain}, achieving truly bidirectional knowledge exchange:

\begin{equation}
\text{Big Data} \ \rightarrow \ \text{Big Model (LLM)} \ \rightarrow \ \text{Big Wisdom (SuperBrain)}.
\end{equation}

Under this mechanism, LLMs would gain AGI-like high-level intelligence through sustained human–machine interaction and accelerated evolution, enabling emergent capabilities in innovation, emotional resonance and human–AI symbiosis.

\subsubsection{Pattern distillation and rule synthesis layer (embedding Meta-LLM Layer into Transformer)}

The integrated architecture of the swarm intelligence framework (Table~\ref{tab:swarm-architecture}) defines four layers:  
\textit{Swarm Layer}, \textit{Fitness Layer}, \textit{Brain Registry} and \textit{Meta-LLM Layer}.  
We propose revisiting the Transformer design to embed these layers directly within the core architecture, creating an ``intelligent Transformer'' optimized for long-term evolution.

In such a design, user-side \textit{Subclass Brains} can execute domain-specific tasks (e.g., UAV vertiport take-off sequence optimization) while contributing success cases and optimization knowledge back to the LLM core.  
For example, in high-dimensional language tasks such as Chinese poetry generation, the system could strengthen its semantic and aesthetic comprehension, mitigating issues such as the ``paradox of poetic intent'' and enhancing language intelligence \citep{weigang2025paradox}.

\paragraph{Potential impact.}
These improvements would not only enhance LLM performance and interpretability but also enable the creation of an open \textit{cognitive ecosystem} where millions of valuable users contribute to long-term human–AI co-evolution.  
Through continuous interaction and wisdom accumulation, this ecosystem could trigger a genuine \textit{intelligence explosion}, accelerating the path toward AGI.  
Although \textit{SuperBrain} remains an early-stage concept, its theoretical grounding and engineering feasibility suggest that--with academic and industrial support--it could mature rapidly in the near future.

\subsection{Once-Learning: A Global Perception Paradigm for SuperBrain}

The concept of \textit{Once-learning} was first introduced in 1998 \citep{weigang1998study} and further elaborated at the IJCNN 1999 conference, where it was applied to meteorological radar image processing \citep{weigang1999study}. This paradigm emphasizes completing holistic perception and feature extraction from a \textbf{single global input} during information processing, rather than relying on multiple, segmented or incremental learning cycles. It is particularly critical in multi-modal information processing and in the treatment of structurally rich languages such as Chinese, where characters inherently possess a two-dimensional structure (\textit{glyph + semantics}) that requires simultaneous perception and integration of both spatial and contextual dimensions \citep{weigang2022new}.

Compared with subsequent related concepts, Once-learning demonstrates clear chronological precedence. The \textit{One-shot learning} approach proposed by Li Fei-Fei et al. in 2003 attracted significant attention in object recognition ~\citep{fe2003bayesian}, eventually inspiring related concepts such as \textit{Zero-shot} and \textit{Few-shot learning}, which have played an important role in the evolution of machine learning, including LLM development \citep{brown2020language}. Similarly, the \textit{YOLO} (\textit{You Only Look Once}) framework proposed by ~\cite{redmon2016you} achieved remarkable engineering success. However, while \textit{Few-shot learning} and \textit{YOLO} both enable efficient learning or detection from limited or single-pass inputs in specific tasks, \textit{Once-learning} was conceived from the outset as a \textbf{cross-domain, general-purpose information processing paradigm}--applicable across neural networks, machine learning and even quantum computing architectures--aligning with the trends of multi-modal processing and multi-dimensional interpretability in AI \citep{althoff2022once, weigang2022new}.

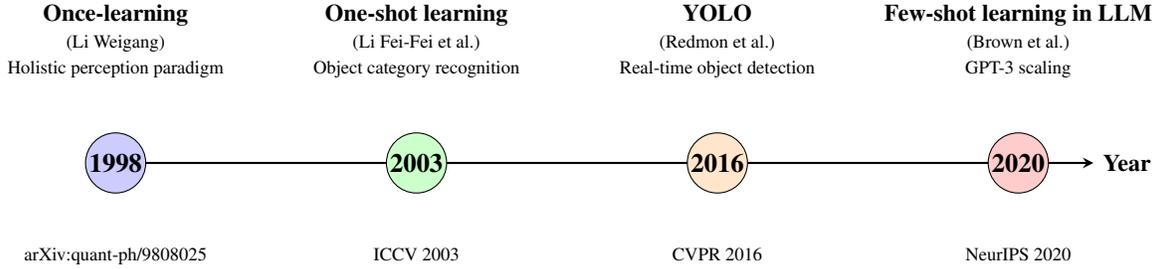
\begin{figure}[htbp]
\centering
\begin{tikzpicture}[
    timeline/.style={thick, ->, >=stealth},
    event/.style={circle, draw, fill=white, minimum size=8mm, inner sep=0pt},
    mylabel/.style={align=center, font=\small}
]

% Draw timeline
\draw[timeline] (0,0) -- (13,0) node[right]{\small \textbf{Year}};

% 1998 - Once-learning
\node[event, fill=blue!20] (e1998) at (0,0) {\textbf{1998}};
\node[mylabel, above=6mm of e1998] {\textbf{Once-learning}\\\scriptsize (Li Weigang)\\\scriptsize Holistic perception paradigm};
\node[mylabel, below=6mm of e1998] {\scriptsize arXiv:quant-ph/9808025};

% 2003 - One-shot learning
\node[event, fill=green!20] (e2003) at (4,0) {\textbf{2003}};
\node[mylabel, above=6mm of e2003] {\textbf{One-shot learning}\\\scriptsize (Li Fei-Fei et al.)\\\scriptsize Object category recognition};
\node[mylabel, below=6mm of e2003] {\scriptsize ICCV 2003};

% 2016 - YOLO
\node[event, fill=orange!20] (e2016) at (8,0) {\textbf{2016}};
\node[mylabel, above=6mm of e2016] {\textbf{YOLO}\\\scriptsize (Redmon et al.)\\\scriptsize Real-time object detection};
\node[mylabel, below=6mm of e2016] {\scriptsize CVPR 2016};

% 2020 - Few-shot in LLM
\node[event, fill=red!20] (e2020) at (12,0) {\textbf{2020}};
\node[mylabel, above=6mm of e2020] {\textbf{Few-shot learning in LLM}\\\scriptsize (Brown et al.)\\\scriptsize GPT-3 scaling};
\node[mylabel, below=6mm of e2020] {\scriptsize NeurIPS 2020};

\end{tikzpicture}
\caption{Timeline of \textbf{Once-learning} and related concepts, highlighting its precedence and conceptual continuity in AI development.}
\label{fig:once_learning_timeline}
\end{figure}

Fig.~\ref{fig:once_learning_timeline} situates the proposed \textit{Once-learning} paradigm within the broader historical context of related AI concepts, highlighting its chronological precedence and conceptual continuity.

In the \textit{SuperBrain} model presented in this work, the integration of Once-learning enhances its viability and impact in three key aspects:
\begin{enumerate}
    \item \textbf{Subclass Brain level:} Improves the capability of each human–LLM cognitive dyad to model global information, including both context and multi-dimensional imagery, ensuring maximum information absorption and pattern extraction within a single interaction, while reducing the information loss inherent to segmented, multi-turn processing.
    \item \textbf{Swarm Intelligence level:} When multiple Subclass Brains share strategies or cognitive signatures, Once-learning enables one-pass fusion of global features from all inputs, as opposed to fragmented feature stacking, thereby preserving richer cross-user collaborative information.
    \item \textbf{Superclass Brain level:} Supports direct holistic pattern matching and reasoning within the global memory module, enabling cross-task and cross-domain cognitive transfer.
\end{enumerate}

Looking forward, Once-learning could inspire \textbf{global tokenization and encoding mechanisms} in future LLM/Transformer architectures, enabling models to perceive and process longer contexts and multi-modal inputs in a single pass. This capability would not only enhance cross-lingual understanding--particularly for high-information-density languages such as Chinese--but also strengthen global constraint modeling in complex engineering tasks such as UAV take-off sequence optimization, 4D flight path conflict resolution and others.

\section{Conclusion}
\label{conclusion}

This paper introduced the \textit{SuperBrain} model, a human–LLM co-evolutionary framework that unifies forward and backward iterative optimization, swarm intelligence and multi-layer cognitive architecture design. By formalizing the concepts of \textit{Subclass Brain} and \textit{Superclass Brain}, the framework shifts from static prompting toward dynamic, interactive and human-centered collective intelligence.

Our experiments on UAV take-off sequence scheduling within an Urban Air Mobility (UAM) scenario illustrate the feasibility of this paradigm. The forward evolutionary process, supported by GA variants (v1–v5) co-developed with LLMs, yielded consistent improvements in responsiveness and worst-case delay, approaching the Round Robin theoretical bound. Importantly, the findings highlight that alignment of the fitness function with operational goals can outweigh raw algorithmic complexity. Moreover, the methodology enables the generation of $(\text{prompt}, f_\lambda, \text{solution})$ triplets, providing the basis for backward optimization and thus completing a closed-loop evolutionary cycle.

\textbf{Contributions.} The work offers three main contributions:  
\begin{enumerate}
 \item A conceptual framework for cognitive co-evolution bridging human–LLM interaction and swarm intelligence;  
 \item An experimental validation in UAM scheduling, showing the practical value of GA-assisted forward evolution;  
 \item A blueprint for integrating forward and backward evolution into a self-improving, multi-layered architecture.
\end{enumerate}

\textbf{Future research} will extend this foundation along three directions:  

\begin{enumerate}
    \item \textbf{Backward Evolution Implementation and Validation} -- Operationalize the theoretical framework of Section~\ref{sec:backward-iter-evo} through controlled experiments, testing $(\text{prompt}, f_\lambda, \text{solution})$ triplets across diverse user–LLM dyads (experts, university students, high school students) to measure variability, reproducibility and robustness. This direction resonates with emerging work on human–LLM collaborative evaluation and iterative prompt optimization \citep{woelfle2024benchmarking}.  

    \item \textbf{Cross-Domain and Cross-Platform Evaluation} -- Apply the SuperBrain framework beyond UAM scheduling, for example in multilingual translation (e.g., poetic intent preservation), energy grid optimization or sensor scheduling, while systematically comparing outcomes across multiple LLM platforms to assess generalizability and alignment stability. This aligns with recent studies on domain transferability and multi-agent LLM ecosystems \citep{park2023generative}.  

    \item \textbf{Architectural and Memory Extensions} -- Investigate embedding long-term user-level memory interfaces, swarm-alignment layers and meta-pattern distillation into Transformer architectures to support persistent cognitive evolution and scalable aggregation into a functional \textit{Superclass Brain}. Related advances include memory-augmented transformers and collective-agent architectures \citep{borgeaud2022improving, liang2023taskmatrix}.  
\end{enumerate}

Ultimately, the vision of \textit{SuperBrain} is to evolve from distributed Subclass Brain interactions toward an emergent Superclass Brain, achieving scalable, explainable and ethically aligned collective intelligence.

%%%%%%%%%%%%%%%%%%%%%%%%%%%%%%%%%%%%%%%%%%%%%%%%%%%%%%%%%%%%%%
\section*{Acknowledgments}
\label{sec:acknowledgments}
We gratefully acknowledge the support of CNPq and the encouragement and insightful feedback from our colleagues and friends. We also note the practical assistance of large language models, including ChatGPT, DeepSeek, Gemini and Grok, which were used to improve writing clarity and technical precision. While the core ideas and interpretations remain the responsibility of the authors, these tools provided valuable support in enhancing the overall productivity of this research.

\bibliographystyle{unsrtnat}
\bibliography{references}

%\section*{Appendix}

\end{document}